\documentclass[10pt,twocolumn,letterpaper]{article}

\usepackage{cvpr}      
\usepackage{multirow}
\definecolor{cvprblue}{rgb}{0.21,0.49,0.74}
\usepackage[pagebackref,breaklinks,colorlinks,allcolors=cvprblue]{hyperref}

\def\paperID{3930} 
\def\confName{CVPR}
\def\confYear{2026}

\title{
Gau-Occ: Geometry-Completed Gaussians for Multi-Modal 3D Occupancy Prediction}

\author{
    Chengxin Lv\textsuperscript{\rm 1,2},
    Yihui Li\textsuperscript{\rm 1,2},
    Hongyu Yang\textsuperscript{\rm 1,3}\thanks{Corresponding author.},
    YunHong Wang\textsuperscript{\rm 2} \\
    \textsuperscript{\rm 1} State Key Laboratory of Virtual Reality Technology and Systems, Beihang University, Beijing, China \\
    \textsuperscript{\rm 2} School of Computer Science and Engineering, Beihang University, Beijing, China\\
    \textsuperscript{\rm 3} School of Artificial Intelligence, Beihang University, Beijing,  China\\
    {\tt\small \{chengxinlv, kidleyh, hongyuyang, yhwang\}@buaa.edu.cn}
}

\begin{document}
\maketitle
\begin{abstract}

3D semantic occupancy prediction is crucial for autonomous driving. While multi-modal fusion improves accuracy over vision-only methods, it typically relies on computationally expensive dense voxel or BEV tensors. We present \textbf{Gau-Occ}, a multi-modal framework that bypasses dense volumetric processing by modeling the scene as a compact collection of semantic 3D Gaussians. To ensure geometric completeness, we propose a \textbf{LiDAR Completion Diffuser (LCD)} that recovers missing structures from sparse LiDAR to initialize robust Gaussian anchors. Furthermore, we introduce \textbf{Gaussian Anchor Fusion (GAF)}, which efficiently integrates multi-view image semantics via geometry-aligned 2D sampling and cross-modal alignment. By refining these compact Gaussian descriptors, Gau-Occ captures both spatial consistency and semantic discriminability. Extensive experiments across challenging benchmarks demonstrate that Gau-Occ achieves state-of-the-art performance with significant computational efficiency.

\end{abstract}    
\section{Introduction}
\label{sec:intro}

3D semantic occupancy prediction is a fundamental capability for autonomous driving, aiming to reconstruct a dense, structured representation of the surrounding 3D environment~\cite{cao_monoscene_2022,miao2023occdepth,huang_tri-perspective_2023,li2025microenhanced}. Early camera-only approaches typically operate on BEV planes~\cite{li2022bevformer,hu2021fiery,philion2020lift} or 3D voxel grids~\cite{garbade2019two,li_voxformer_2023}.
However, their performance is limited by weak geometric cues, especially in distant or occluded regions. This limitation often leads to incomplete occupancy estimates and coarse free-space predictions in complex driving scenes.

To address these limitations, recent works integrate active depth sensors such as LiDAR or radar with multi-view RGB~\cite{pan_co-occ_2024,wang_occgen_2025,li_occmamba_2025}, exploiting complementary geometric and semantic information. Despite notable progress, two main challenges remain: \textit{(i)} raw point clouds are sparse and occlusion-biased, capturing mostly visible surfaces while missing many occupied but unobserved regions, limiting the completeness of 3D reasoning; \textit{(ii)} mainstream fusion pipelines are computationally heavy. Early-fusion schemes either project points into multiple image views~\cite{vora2020pointpainting} or lift dense image features into volumetric grids~\cite{sindagi2019mvx}, while transformer-based fusion in voxel or BEV space incurs prohibitive memory and computation~\cite{pan_co-occ_2024,BEVFusion}, thus hindering scalability to higher spatial resolution or longer temporal horizons.

We advocate a compact and unified 3D representation that preserves geometric fidelity while enabling effective cross-modal fusion. Recent advances in 3D Gaussian ~\cite{li2025micro,li2026tokensplat} primitives demonstrate that such representations can model scene geometry and semantics with high expressiveness from multi-view observations~\cite{3dgs,chen2026catalyst4dhighfidelity3dto4dscene,huang_gaussianformer_2025}. While promising, existing Gaussian-based approaches are predominantly vision-only, and their application to multi-modal occupancy prediction remains underexplored, particularly under real-world constraints such as sparse LiDAR sampling and limited computational budgets.

We propose \textbf{Gau-Occ}, a framework that leverages learnable semantic Gaussian anchors for efficient scene representation.
Initialized from completed LiDAR scans, these anchors are iteratively refined in a feed-forward manner by selectively fusing multi-view image features.
The refined anchors are then splatted into voxel space, and their semantic contributions are accumulated to generate the final 3D occupancy predictions, all while maintaining computational efficiency and avoiding dense voxel costs. We instantiate this pipeline through two dedicated components:

First, the \textit{\textbf{LiDAR Completion Diffuser (LCD)}} reconstructs dense, geometrically consistent points from sparse, occlusion-biased LiDAR scans. 
Rather than merely increasing point density, LCD learns structural priors from aggregated LiDAR sweeps, capturing the continuity of surfaces and the regularity of structures, allowing it to infer plausible and metrically aligned geometry in unobserved or heavily occluded regions. 
This produces geometry-faithful anchors for subsequent Gaussian-based reasoning.

Second, we propose the \textit{\textbf{Gaussian Anchor Fusion (GAF)}} module, which aligns multi-view image semantics with a LiDAR-anchored 3D structural prior. Each Gaussian anchor reprojects onto image planes and performs local feature sampling through adaptive 2D offsets, conditioned on its LiDAR feature. A geometry-aware VLAD (Vector of Locally Aggregated Descriptors)  mechanism~\cite{jegou2010aggregating} then aggregates the sampled image features into compact, view-consistent descriptors. These descriptors are modulated by the anchor features and fused via a single cross-attention layer.  As a result, GAF effectively bridges the dense semantic richness of images with the precise geometry of LiDAR, yielding a deeply integrated representation for robust 3D occupancy prediction. By operating solely over anchor points, GAF maintains spatial precision while significantly reducing computational overhead.

In summary, our contributions are:
\begin{itemize}
\item We propose Gau-Occ, a compact Gaussian-based framework that unifies LiDAR and multi-view images for 3D semantic occupancy prediction.
\item We introduce LCD, a learned module that enhances geometric completeness under sparse depth sampling.
\item We present GAF, a geometry-aligned fusion module that aggregates multi-view image features into Gaussian anchors efficiently and accurately.
\end{itemize}


\section{Related Work}

\subsection{Semantic Occupancy Prediction}
3D semantic occupancy prediction has become a key paradigm for dense environment modeling in perception. Unlike detection or instance/semantic segmentation that target discrete entities, occupancy estimation provides \emph{voxel-level} geometric and semantic labels, enabling fine-grained understanding of both static layouts and dynamic objects. Early efforts primarily addressed indoor scenes~\cite{song_semantic_2017, li_anisotropic_2020}, while recent advances extend to outdoor driving with LiDAR-, camera-, or hybrid-based inputs~\cite{cao_monoscene_2022, miao2023occdepth, huang_tri-perspective_2023,zhou2025autoocc}. 

A central challenge is the choice of a 3D scene representation. Voxel-based methods~\cite{Zhou_2018_CVPR, s18103337} can capture fine details but incur heavy memory/compute due to dense volumetric tensors. To reduce redundancy, many approaches leverage 2D planar approximations such as BEV~\cite{li2022bevformer, Yang_2023_CVPR} or tri-plane views~\cite{huang_tri-perspective_2023}; however, dimensional collapse inevitably discards spatial detail and introduces aliasing across depth. In contrast, semantic Gaussian representations~\cite{huang_gaussianformer_2025,GaussianOcc_2025_ICCV,zuo2025gaussianworld,GaussRender_2025_ICCV} encode only non-empty regions via a set of learnable 3D Gaussians, providing a compact yet expressive modeling of geometry and semantics.
Recognizing that robust autonomy requires multi-modal sensing, we repurpose these Gaussian primitives as a unified anchor representation for fusing LiDAR and camera data, thereby retaining spatial fidelity while avoiding
dense volumetric computation.

\subsection{LiDAR-Camera Fusion in 3D Perception}
Fusing complementary LiDAR and camera data is pivotal for robust 3D occupancy prediction. Existing strategies can be grouped into three families. 
\textbf{Projection-level fusion}~\cite{huang_epnet_2020, MSeg3D} projects LiDAR points to image planes or lifts pixels into 3D; it is simple and efficient but sensitive to calibration errors and viewpoint mismatch. 
\textbf{Feature-level fusion}~\cite{Zhang_2023_ICCV, BEVFusion} independently encodes each modality and aggregates features in a shared voxel or BEV space; while effective, this typically introduces substantial memory/runtime overhead in dense 3D spaces. 
\textbf{Attention-based fusion}~\cite{TransFusion, DeepFusion, zhao2025gaussianformer3dmultimodalgaussianbasedsemantic} employs cross-attention to learn adaptive correspondences without strict geometric alignment, but can remain costly when performed over large voxel/BEV tensors. In this work, we propose the \emph{GAF} module, where learnable 3D Gaussians act as spatially aware queries to aggregate multi-view image features. 


\definecolor{others}{cmyk}{0,0,0,1}  
\definecolor{barrier}{cmyk}{0.0,0.53,0.80,0.0}            
\definecolor{bicycle}{cmyk}{0.0,0.25,0.20,0.0}            
\definecolor{bus}{cmyk}{0.0,0.0,1.0,0.0}                  
\definecolor{car}{cmyk}{1.0,0.39,0.0,0.04}                
\definecolor{constveh}{cmyk}{1.0,0.0,0.0,0.0}             
\definecolor{motorcycle}{cmyk}{0.0,0.50,1.0,0.0}          
\definecolor{pedestrian}{cmyk}{0.0,1.0,1.0,0.0}           
\definecolor{trafficcone}{cmyk}{0.0,0.06,0.41,0.0}        
\definecolor{trailer}{cmyk}{0.0,0.56,1.0,0.47}            
\definecolor{truck}{cmyk}{0.33,0.87,0.0,0.06}             
\definecolor{drivesurf}{cmyk}{0.0,1.0,0.0,0.0}            
\definecolor{otherflat}{cmyk}{0.0,0.01,0.01,0.45}         
\definecolor{sidewalk}{cmyk}{0.0,1.0,0.0,0.71}            
\definecolor{terrain}{cmyk}{0.38,0.0,0.67,0.06}           
\definecolor{manmade}{cmyk}{0.08,0.08,0.0,0.02}           
\definecolor{vegetation}{cmyk}{1.0,0.0,1.0,0.31}          

\begin{figure*}[t]
\centering
\includegraphics[width=0.96\textwidth]{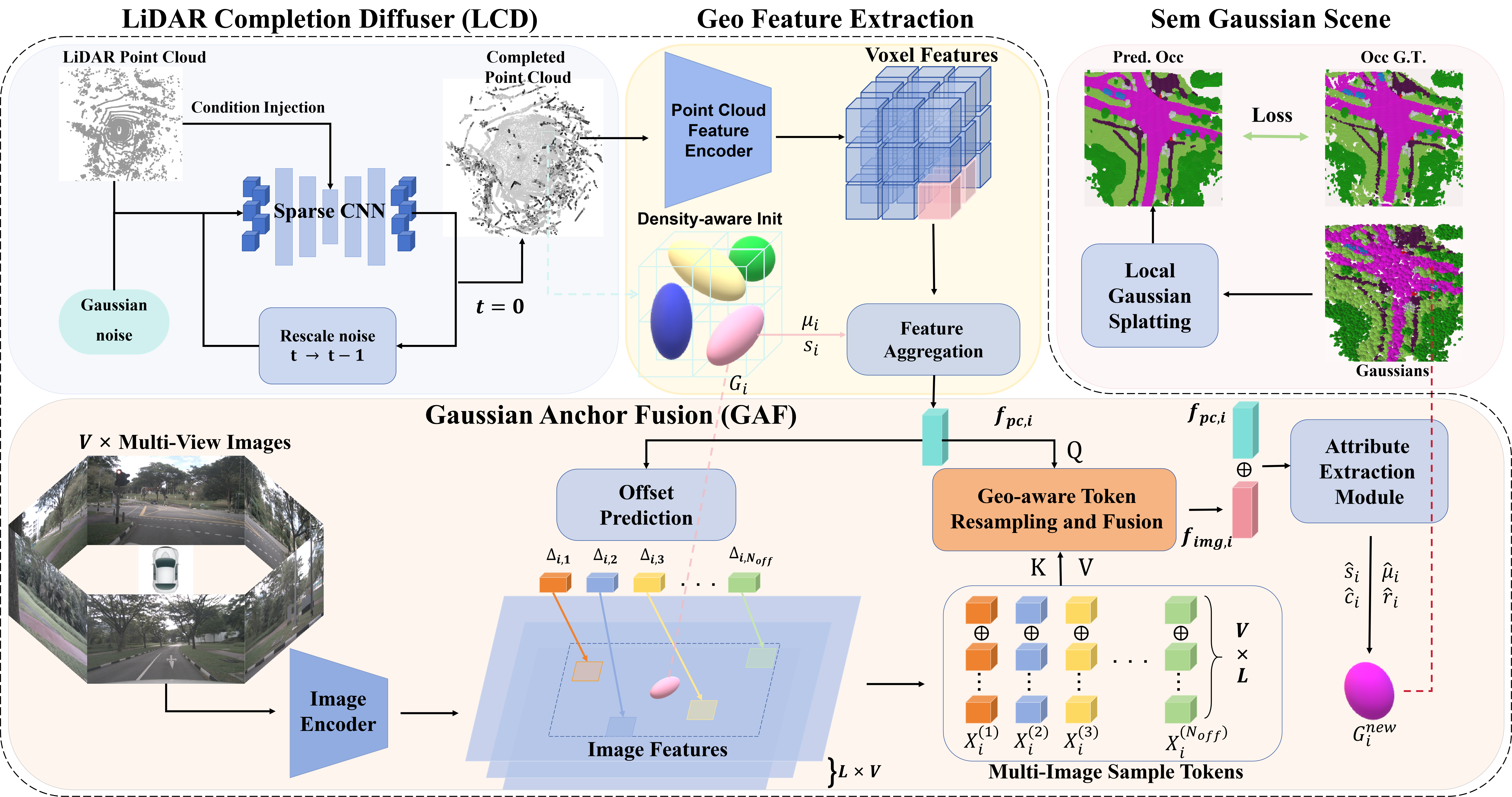}
\caption{
Overview of \textbf{Gau-Occ}.
Sparse LiDAR scans are first completed by a pretrained \textbf{LiDAR Completion Diffuser (LCD)} to recover occluded geometry. The completed points are encoded into geometric features to initialize density-aware semantic 3D Gaussians. Each Gaussian then anchors multi-view image features via our \textbf{Gaussian Anchor Fusion (GAF)}, producing geometry-aligned multi-modal representations. The refined Gaussians are finally splatted into voxel space for semantic occupancy prediction.
}
\label{pipeline}
\end{figure*}
\section{Proposed Approach}
We propose \textbf{Gau-Occ}, a compact representation of 3D scenes using semantic Gaussians that jointly encode LiDAR geometry and multi-view semantics. As shown in Fig.~\ref{pipeline}, sparse LiDAR scans are first completed by a \textit{LiDAR Completion Diffuser (LCD)} to recover occluded or unobserved structures. The completed points are then voxelized into sparse features that initialize density-aware Gaussians. Each Gaussian aggregates visual cues through the proposed \textit{Gaussian Anchor Fusion (GAF)}, which predicts geometry-guided sampling offsets and performs cross-modal feature refinement via a VLAD-style descriptor. These descriptors are modulated by the geometry features and fused via a single cross-attention layer. This produces view-consistent, semantically discriminative anchor features that update Gaussian attributes. Finally, the refined Gaussians are splatted into voxel space to generate dense 3D semantic occupancy.

\subsection{3D Semantic Gaussian Scene Representation}
Semantic occupancy prediction aims to jointly infer geometry and semantic labels in 3D space.  
Given a sparse LiDAR point cloud $\mathcal{P}=\{P_i\in\mathbb{R}^3\}_{i=1}^{N_P}$ and multi-view images $\mathcal{I}=\{I_j\in\mathbb{R}^{3\times H\times W}\}_{j=1}^{N_I}$, the task is to predict a voxelized semantic occupancy grid $O\in\mathbb{R}^{|\mathcal{C}|\times X\times Y\times Z}$, where $|\mathcal{C}|$ is the number of semantic classes and $(X,Y,Z)$ defines the voxel resolution.

We model the scene as a set of semantic 3D Gaussians $\mathcal{G}=\{G_i\}_{i=1}^{N_G}$, where each $G_i$ is parameterized by center $\boldsymbol{\mu}\in\mathbb{R}^3$, rotation quaternion $\mathbf{r}\in\mathbb{R}^4$, scale $\mathbf{s}\in\mathbb{R}^3$, and semantic vector $\mathbf{c}\in\mathbb{R}^{|\mathcal{C}|}$.  
The semantic contribution of a Gaussian at a query position $\mathbf{x}\in\mathbb{R}^3$ is defined as:
\begin{equation}
\mathbf{g}(\mathbf{x}; G_i) = \exp\Big(-\tfrac{1}{2} (\mathbf{x}-\boldsymbol{\mu})^\top \boldsymbol{\Sigma}^{-1} (\mathbf{x}-\boldsymbol{\mu}) \Big) \cdot \mathbf{c},
\end{equation}
where the covariance matrix is: 
\begin{equation}
\boldsymbol{\Sigma} = \mathbf{R}\mathbf{S}\mathbf{S}^\top\mathbf{R}^\top,\quad
\mathbf{S} = \operatorname{diag}(\mathbf{s}),\quad
\mathbf{R} = \operatorname{q2r}(\mathbf{r}),
\end{equation}
and $\operatorname{q2r}(\cdot)$ converts a unit quaternion to a rotation matrix.

The predicted occupancy at $\mathbf{x}$ is obtained by aggregating contributions from all Gaussians:
\begin{equation}
\hat{\mathbf{o}}(\mathbf{x}) = \sum_{G_i \in \mathcal{G}} \mathbf{g}(\mathbf{x}; G_i).
\end{equation}

To ensure efficiency, we adopt \textit{local Gaussian splatting}, where each voxel only aggregates Gaussians within its spatial neighborhood, preserving spatial precision while avoiding full-scene accumulation.

Following \cite{huang_tri-perspective_2023}, we optimize the model with a joint objective $\mathcal{L}_{\text{CE}}+\mathcal{L}_{\text{Lov}}$, combining cross-entropy and Lovász-Softmax losses to enhance segmentation accuracy and class balance.
The sparse and fully differentiable formulation preserves fine geometric details while maintaining efficient aggregation and gradient propagation, providing a unified foundation for subsequent multi-modal fusion.

\subsection{LiDAR Completion Diffuser (LCD)}

Outdoor LiDAR scans are sparse and occlusion-biased due to limited angular resolution and visibility. We propose the LiDAR Completion Diffuser (LCD), a local diffusion model that reconstructs dense, geometrically consistent point clouds from sparse scans. 
Unlike conventional  DDPMs~\cite{ho2020denoising}, which apply global noise and scaling that may distort metric geometry, LCD performs point-wise local diffusion. By perturbing each 3D point independently within its local neighborhood, it strictly preserves absolute scale and fine details.

Given a raw LiDAR scan $\mathcal{P}=\{P_i\in\mathbb{R}^3\}_{i=1}^{N_P}$, the completion objective is to generate a densified point cloud $\mathcal{P}'=\{P'_i\}_{i=1}^{N_{P'}}$ that approximates a dense supervision target $\mathcal{T}=\{\mathcal{T}_j\in\mathbb{R}^3\}_{j=1}^{N_T}$.  
Following common practice in LiDAR self-supervision~\cite{pan_co-occ_2024,SDGOccDuan_2025_CVPR}, $\mathcal{T}$ is constructed by aggregating $K$ temporally adjacent, ego-motion–aligned sweeps from the same scene, providing dense ground-truth geometry for training while maintaining scene-level consistency.
\noindent\textbf{Forward Process.}  
Each ground-truth point $\mathcal{T}_j$ is locally perturbed:
\begin{equation}
\mathcal{T}_j^{(t)} = \,\mathcal{T}_j + \sqrt{1-\bar{\alpha}_t}\,\boldsymbol{\epsilon}, 
\quad \boldsymbol{\epsilon}\sim\mathcal{N}(\mathbf{0},\mathbf{I}),
\end{equation}
where $\bar{\alpha}_t$ follows a linear noise schedule from $t=1$ to $T$, as in DDPM~\cite{ho2020denoising}, and no global scaling term is applied to preserve the scene's metric structure.

\noindent\textbf{Reverse Process.}  
The denoising network $\hat{\boldsymbol{\epsilon}}_{\theta}$ learns to predict the injected noise  conditioned on the sparse input $\mathcal{P}$:
\begin{equation}
\mathcal{L}_{\text{diff}} = \big\|\boldsymbol{\epsilon} - \hat{\boldsymbol{\epsilon}}_{\theta}(\mathcal{T}^{(t)}, \mathcal{P}, t)\big\|_2^2,
\end{equation}
where $\mathcal{T}^{(t)}=\{\mathcal{T}_j^{(t)}\}$ denotes the perturbed points at timestep $t$.  
Through iterative denoising, LCD reconstructs the clean target $\mathcal{T}$ conditioned on the sparse $\mathcal{P}$, effectively learning spatial priors for occluded and unobserved regions.


\subsection{Gaussian Initialization from Completed LiDAR}
Given the completed LiDAR cloud $\mathcal{P}'$ from LCD, we initialize a compact set of semantic 3D Gaussians that ensures both comprehensive geometric coverage and structural diversity. Unlike GaussianFormer~\cite{huang_gaussianformer_2025}, which uses random sampling, we employ a hybrid geometry-aware initialization, balancing density and coverage.

\noindent\textbf{Density-based Selection.}  For each point $P'_i$, local density within a radius $R_d$ is estimated, and the highest-density locations are iteratively chosen as Gaussian centers $\boldsymbol{\mu}_{d_i}$. Neighbors within $R_d$ are suppressed to prevent redundancy. This continues until $N_d$ centers $\mathcal{P}_d$ are obtained, capturing detailed and frequently observed surfaces.

\noindent\textbf{Random Coverage Sampling.}  
From the remaining points, $N_r$ centers $\mathcal{P}_r$ are uniformly sampled to cover sparse or low-texture regions. 
The union of both subsets forms the initialization set:
\begin{equation}
\mathcal{P}_{\text{init}} = \mathcal{P}_d \cup \mathcal{P}_r.
\end{equation}

\begin{figure}[t]
\centering
\includegraphics[width=0.9\columnwidth]{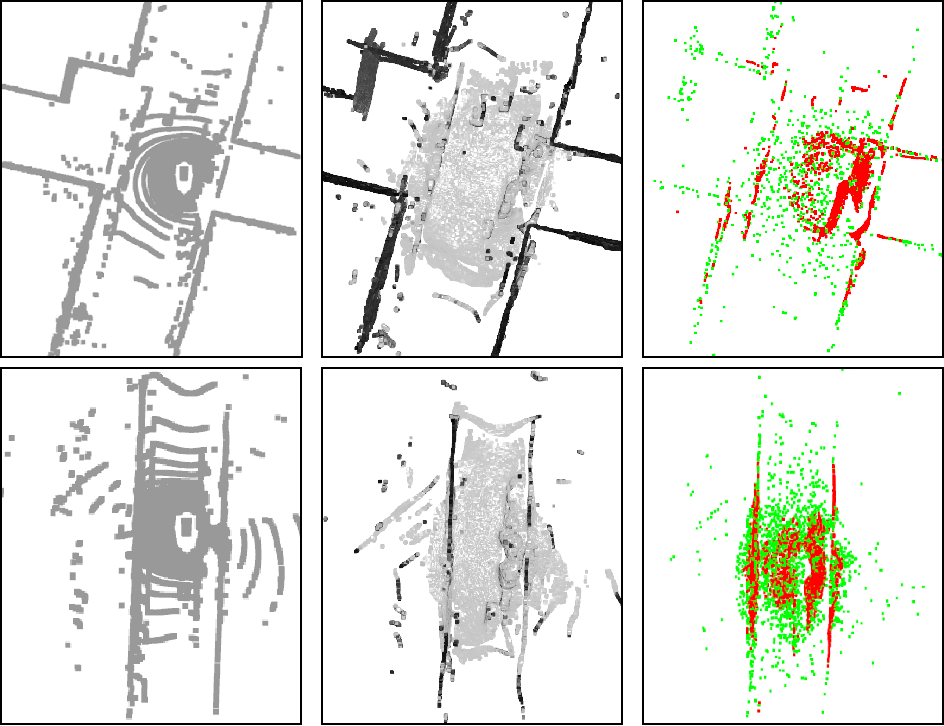} 
\caption{
Hybrid Gaussian initialization. 
Left: raw sparse LiDAR input. 
Middle: completed point cloud $\mathcal{P}'$ from LCD. 
Right: initialized Gaussian centers derived from $\mathcal{P}'$, with density-based subset $\mathcal{P}_d$ (red) and random subset $\mathcal{P}_r$ (green).
}
\label{fig4}
\end{figure}

Each center $\boldsymbol{\mu}_i \in \mathcal{P}_{\text{init}}$ is assigned an axis-aligned initial scale $\mathbf{s}_i = (s_x, s_y, s_z)$, forming the Gaussian set
$\mathcal{G} = \{{G_i\} = (\boldsymbol{\mu}_i, \mathbf{s}_i)}_{i=1}^{N_G}$.
As illustrated in Fig.~\ref{fig4}, this hybrid initialization produces spatially balanced, geometry-aligned Gaussians, providing a robust foundation for the subsequent multi-modal feature fusion.

\subsection{Gaussian Anchor Fusion (GAF)}
To unify precise LiDAR geometry with rich image semantics, we propose Gaussian Anchor Fusion (GAF), a geometry-conditioned multi-modal fusion module that extracts, samples, and aggregates features for each Gaussian anchor. With the initialized set $\mathcal{G} = \{{G_i}\}$, each anchor acts as a 3D query linking LiDAR and image domains.

\noindent\textbf{Geometry Feature Extraction.}
The completed LiDAR cloud $\mathcal{P}'$ is voxelized into a sparse grid of size $D\times H\times W$, keeping at most $T_p=10$ points per voxel~\cite{yan2018second}. For each occupied voxel $v\in\mathcal{V}$, we average point-wise embeddings $\psi(p)$ (coordinates, intensity, etc.) to form $\mathbf{f}^0_v$ and feed into a 3D sparse CNN to obtain voxel-wise features $\mathbf{F}_v$.

For a Gaussian anchor centered at $\boldsymbol{\mu}_i$ with scale $\mathbf{s}_i=(s_x,s_y,s_z)$, the adaptive neighborhood radius is:
\begin{equation}
R_{\mathrm{geo}} = k\,\overline{s}_i = \tfrac{k}{3}(s_x+s_y+s_z),
\end{equation}
where $k$ is a constant controlling context range.  
Neighboring voxel features within $\mathcal{N}(\boldsymbol{\mu}_i, R_{\mathrm{geo}})$ are aggregated via an exponential distance kernel:
\begin{equation}
\mathbf{f}_{\mathrm{pc},i} =
\frac{\sum_{v\in\mathcal{N}(\boldsymbol{\mu}_i)}w_v\,\mathbf{F}_v}
{\sum_{v\in\mathcal{N}(\boldsymbol{\mu}_i)}w_v},
\quad
w_v = \exp(-\gamma\|\mathbf{p}_v-\boldsymbol{\mu}_i\|_2),
\end{equation}
where $\mathbf{p}_v$ denotes the center of voxel $v$ and $\gamma$ is a fall-off coefficient. 
This yields the geometry-aware anchor descriptor $\mathbf{f}{_\mathrm{pc},_i}\in\mathbb{R}^{d_{pc}}$.

\noindent\textbf{Geometry-guided Image Sampling.}
Multi-scale image features $\mathbf{F}_v^{(l)} = g_l(I_v)$ are extracted using ResNet-50 with FPN,
where $g_l(\cdot)$ denotes the feature extractor at level $l$ and $v$ indicates camera view.
Each Gaussian center $\boldsymbol{\mu}_i$ is projected to the $v$-th camera via the differentiable projection function $\Pi_v:\mathbb{R}^3\!\rightarrow\!\mathbb{R}^2$, yielding the reference pixel:
\begin{equation}
\mathbf{pix}_{i,v} = \Pi_v(\boldsymbol{\mu}_i).
\end{equation}

At level $l$, we sample a small local region around $\mathbf{pix}_{i,v}$ by predicting $N_{\text{off}}$ normalized 2D offsets $\Delta_{i,r}\!\in\!(-1,1)^2$ with a two-layer MLP conditioned on $\mathbf{f}_{\mathrm{pc},i}$:
\begin{equation}
\mathbf{x}^{(r)}_{i,v,l} = \frac{\mathbf{pix}_{i,v}}{s_l} + \Delta_{i,r}\,R_l,
\quad r = 1,\dots,N_{\text{off}},
\end{equation}
where $s_l$ is the downsampling stride of FPN level $l$, and $R_l$ specifies the sampling radius (in feature-map pixels).  
Conditioning the offsets on $\mathbf{f}_{\mathrm{pc},i}$ aligns the sampling process with underlying scene geometry, improving spatial consistency and long-range correspondence across views.

\begin{figure}[t]
\centering
\includegraphics[width=0.99\linewidth]{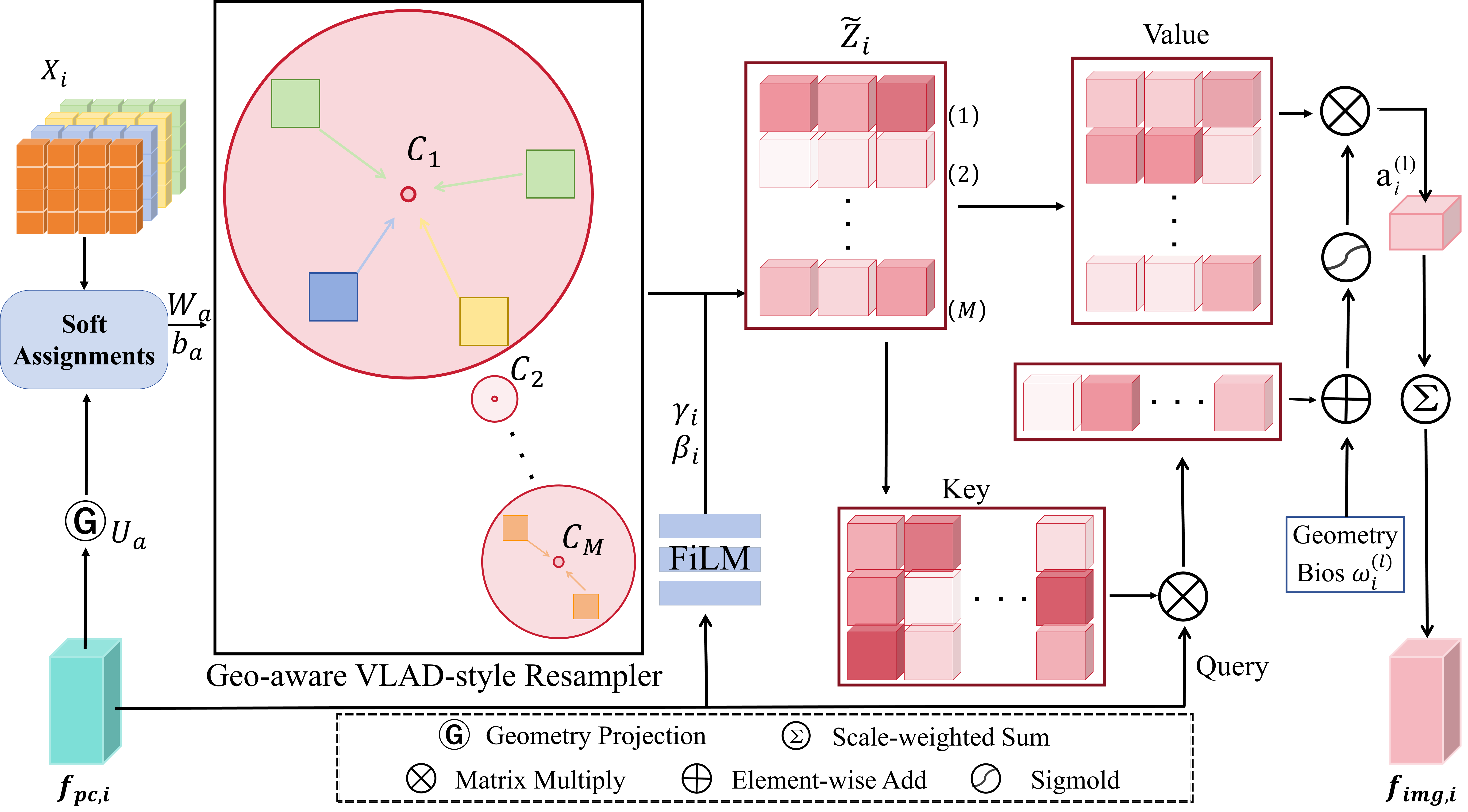} 
\caption{
Schematic of geometry-aware image token resampling and modulation.
}
\label{Geo-VLAD}
\end{figure}

\noindent\textbf{Geometry-aware Token Resampling and Fusion.}
From all views and pyramid levels, we bilinearly sample feature tokens at $\mathbf{x}^{(r)}_{i,v,l}$ and stack them into $\mathbf{X}_i\!\in\!\mathbb{R}^{N\times d}$, where $N=V\times L\times N{_\text{off}}$ is the total number of samples. 
As shown in Fig.~\ref{Geo-VLAD}, instead of applying another attention block, we aggregate them via a geometry-aware VLAD-style~\cite{jegou2010aggregating} resampler using codewords $\{\mathbf{C}_m\}_{m=1}^{M}$ that act as learnable semantic prototypes in feature space:
\begin{equation}
\alpha_{i,n,m} = \operatorname{softmax}_{m}\!\big([W_a\mathbf{x}_{i,n}]_m + [U_a\mathbf{f}_{\mathrm{pc},i}]_m + b_m\big),
\end{equation}
\begin{equation}
\mathbf{Z}_i = \operatorname{stack}_{m=1}^{M}\!\Big(
W_z\,\operatorname{normalize}\!\big(\sum_{n=1}^{N}\alpha_{i,n,m}(\mathbf{x}_{i,n}-\mathbf{C}_m)\big)
\Big),
\end{equation}
where $\mathbf{x}_{i,n}$ is the $n$-th sampled token, $\operatorname{normalize}(\cdot)$ performs $\ell_2$ normalization, and $W_a,U_a,W_z$ are learnable linear projections. By conditioning the assignment $\alpha_{i,n,m}$ on the LiDAR feature, our aggregation becomes geometry-aware.
FiLM modulation~\cite{perez2018film} further rescales and shifts features, enabling a more adaptive fusion:
\begin{equation}
\gamma_i,\beta_i = \operatorname{MLP}_{\text{FiLM}}(\mathbf{f}_{\mathrm{pc},i}),
\quad
\widetilde{\mathbf{Z}}_i = \gamma_i\odot\mathbf{Z}_i + \beta_i,
\end{equation}
where $\odot$ denotes element-wise multiplication.
Cross-attention is then performed between the LiDAR anchor (as query) and the modulated visual tokens (as keys/values):
\begin{equation}
\mathbf{a}_i^{(l)} = 
\operatorname{softmax}\!\left(\frac{\mathbf{Q}_i(\mathbf{K}_i^{(l)})^{\!\top}}{\sqrt{d}} + \log w_i^{(l)}\right)\mathbf{V}_i^{(l)},
\end{equation}
where $\mathbf{Q}_i=W_q\mathbf{f}_{\mathrm{pc},i}$, 
$\mathbf{K}_i^{(l)}=\widetilde{\mathbf{Z}}_iW_k^{(l)}$,
$\mathbf{V}_i^{(l)}=\widetilde{\mathbf{Z}}_iW_v^{(l)}$, 
and $w_i^{(l)}$ is a spatial weight encoding reprojection consistency:
\begin{equation}
w_i^{(l)}=\exp\!\left(-\frac{\|\mathbf{pix}_{i,v}-\Pi_v(\boldsymbol{\mu}_i)\|^2}{2\sigma_l^2}\right),
\quad \sigma_l=\kappa\,R_l,
\end{equation}
where $\kappa$ is a scalar bandwidth coefficient that ties $\sigma_l$ to sampling radius $R_l$. The multi-scale aggregated descriptor is:
\begin{equation}
\mathbf{f}_{\mathrm{img},i} = \sum_{l=1}^{L}\lambda_l\,\mathbf{a}_i^{(l)},
\end{equation}
where $\lambda_l$ are learnable scale weights. This entire pipeline results in a spatially precise and semantically rich representation for occupancy prediction.

Finally, fused features $[\mathbf{f}_{\mathrm{pc},i};\mathbf{f}_{\mathrm{img},i}]$ are decoded through a two-layer FFN to update Gaussian attributes:
\begin{equation}
[\widehat{\boldsymbol{\mu}}_i,\widehat{\mathbf{s}}_i,\widehat{\mathbf{r}}_i,\widehat{\mathbf{c}}_i]
=\operatorname{FFN}\!\big([\mathbf{f}_{\mathrm{pc},i};\mathbf{f}_{\mathrm{img},i}]\big),
\end{equation}
The refined Gaussian
$
\mathbf{G}_i^{\text{new}}=(\boldsymbol{\mu}_i+\widehat{\boldsymbol{\mu}}_i,\ \widehat{\mathbf{s}}_i,\ \widehat{\mathbf{r}}_i,\ \widehat{\mathbf{c}}_i)
$
is splatted  to produce semantic occupancy prediction $O$.

\section{Experiments}

\begin{table*}[t]
\caption{Quantitative comparison on \textbf{SurroundOcc-nuScenes} validation set. The best results are in \textbf{bold}, second best are \underline{underlined}. 
}
\centering

\resizebox{\textwidth}{!}{%
\begin{tabular}{c|c|cc|cccccccccccccccc}
\toprule
\multirow{3}{*}[-0.5em]{Method} & \multirow{3}{*}[-0.5em]{Modality} & \multirow{3}{*}[-0.5em]{\makebox[1cm]{IoU$\uparrow$}} & \multirow{3}{*}[-0.5em]{\makebox[1cm]{mIoU$\uparrow$}} & \makebox[0.8cm]{\rotatebox{90}{barrier}} & \makebox[0.8cm]{\rotatebox{90}{bicycle}} & \makebox[0.8cm]{\rotatebox{90}{bus}} & \makebox[0.8cm]{\rotatebox{90}{car}} & \makebox[0.8cm]{\rotatebox{90}{const. veh.}} & \makebox[0.8cm]{\rotatebox{90}{motorcycle}} & \makebox[0.8cm]{\rotatebox{90}{pedestrian}} & \makebox[0.8cm]{\rotatebox{90}{traffic cone}} & \makebox[0.8cm]{\rotatebox{90}{trailer}} & \makebox[0.8cm]{\rotatebox{90}{truck}} & \makebox[0.8cm]{\rotatebox{90}{drive. surf.}} & \makebox[0.8cm]{\rotatebox{90}{other flat}} & \makebox[0.8cm]{\rotatebox{90}{sidewalk}} & \makebox[0.8cm]{\rotatebox{90}{terrain}} & \makebox[0.8cm]{\rotatebox{90}{manmade}} & \makebox[0.8cm]{\rotatebox{90}{vegetation}} \\
& & & & \makebox[0.8cm]{\colorbox{barrier}{\rule{0pt}{4pt}\rule{4pt}{0pt}}} & \makebox[0.8cm]{\colorbox{bicycle}{\rule{0pt}{4pt}\rule{4pt}{0pt}}} & \makebox[0.8cm]{\colorbox{bus}{\rule{0pt}{4pt}\rule{4pt}{0pt}}} & \makebox[0.8cm]{\colorbox{car}{\rule{0pt}{4pt}\rule{4pt}{0pt}}} & \makebox[0.8cm]{\colorbox{constveh}{\rule{0pt}{4pt}\rule{4pt}{0pt}}} & \makebox[0.8cm]{\colorbox{motorcycle}{\rule{0pt}{4pt}\rule{4pt}{0pt}}} & \makebox[0.8cm]{\colorbox{pedestrian}{\rule{0pt}{4pt}\rule{4pt}{0pt}}} & \makebox[0.8cm]{\colorbox{trafficcone}{\rule{0pt}{4pt}\rule{4pt}{0pt}}} & \makebox[0.8cm]{\colorbox{trailer}{\rule{0pt}{4pt}\rule{4pt}{0pt}}} & \makebox[0.8cm]{\colorbox{truck}{\rule{0pt}{4pt}\rule{4pt}{0pt}}} & \makebox[0.8cm]{\colorbox{drivesurf}{\rule{0pt}{4pt}\rule{4pt}{0pt}}} & \makebox[0.8cm]{\colorbox{otherflat}{\rule{0pt}{4pt}\rule{4pt}{0pt}}} & \makebox[0.8cm]{\colorbox{sidewalk}{\rule{0pt}{4pt}\rule{4pt}{0pt}}} & \makebox[0.8cm]{\colorbox{terrain}{\rule{0pt}{4pt}\rule{4pt}{0pt}}} & \makebox[0.8cm]{\colorbox{manmade}{\rule{0pt}{4pt}\rule{4pt}{0pt}}} & \makebox[0.8cm]{\colorbox{vegetation}{\rule{0pt}{4pt}\rule{4pt}{0pt}}} \\
& & & & & & & & & & & & & & & & & & & \\
\midrule
MonoScene\cite{cao_monoscene_2022} & C & 24.0 & 7.3 & 4.0 & 0.4 & 8.0 & 8.0 & 2.9 & 0.3 & 1.2 & 0.7 & 4.0 & 4.4 & 27.7 & 5.2 & 15.1 & 11.3 & 9.0 & 14.9 \\
C-CONet\cite{wang2023openoccupancy} & C & 26.1 & 18.4 & 18.6 & 10.0 & 26.4 & 27.4 & 8.6 & 15.7 & 13.3 & 9.7 & 10.9 & 20.2 & 33.0 & 20.7 & 21.4 & 21.8 & 14.7 & 21.3 \\
GaussianFormer\cite{huang_gaussianformer_2025} & C & 29.8 & 19.1 & 19.5 & 11.3 & 26.1 & 29.8 & 10.5 & 13.8 & 12.6 & 8.7 & 12.7 & 21.6 & 39.6 & 23.3 & 24.5 & 23.0 & 9.6 & 19.1 \\
BEVFormer\cite{li2022bevformer} & C & 30.5 & 16.8 & 14.2 & 6.6 & 23.5 & 28.3 & 8.7 & 10.8 & 6.6 & 4.1 & 11.2 & 17.8 & 37.3 & 18.0 & 22.9 & 22.2 & 13.8 & 22.2 \\
TPVFormer\cite{huang_tri-perspective_2023} & C & 30.9 & 17.1 & 16.0 & 5.3 & 23.9 & 27.3 & 9.8 & 8.7 & 7.1 & 5.2 & 11.0 & 19.2 & 38.9 & 21.3 & 24.3 & 23.2 & 11.7 & 20.8 \\
OccFormer\cite{zhang_occformer_2023} & C & 31.4 & 19.0 & 18.7 & 10.4 & 23.9 & 30.3 & 10.3 & 14.2 & 13.6 & 10.1 & 12.5 & 20.8 & 38.8 & 19.8 & 24.2 & 22.2 & 13.5 & 21.4 \\
SurroundOcc\cite{wei_surroundocc_2023} & C & 31.5 & 20.3 & 20.6 & 11.7 & 28.1 & 30.9 & 10.7 & 15.1 & 14.1 & 12.1 & 14.4 & 22.3 & 37.3 & 23.7 & 26.5 & 22.8 & 14.9 & 21.9 \\

FB-Occ\cite{li2023fbocc} & C & 31.5 & 19.6 & 20.6 & 11.3 & 26.9 & 29.8 & 10.4 & 13.6 & 13.7 & 11.4 & 11.5 & 20.6 & 38.2 & 21.5 & 24.6 & 22.7 & 14.8 & 21.6 \\
GaussianFormer-2\cite{huang_gaussianformer-2_2025} & C & 31.7 & 20.8 & 21.4 & 13.4 & 28.5 & 30.8 & 10.9 & 15.8 & 13.6 & 10.5 & 14.0 & 22.9 & 40.6 & 24.4 & 26.1 & 24.3 & 13.8 & 22.0 \\
\midrule
LMSCNet\cite{roldao2020lmscnet} & L & 36.6 & 14.9 & 13.1 & 4.5 & 14.7 & 22.1 & 12.6 & 4.2 & 7.2 & 7.1 & 12.2 & 11.5 & 26.3 & 14.3 & 21.1 & 15.2 & 18.5 & 34.2 \\
L-CONet\cite{wang2023openoccupancy} & L & 39.4 & 17.7 & 19.2 & 4.0 & 15.1 & 26.9 & 6.2 & 3.8 & 6.8 & 6.0 & 14.1 & 13.1 & 39.7 & 19.1 & 24.0 & 23.9 & 25.1 & 35.7 \\
\midrule
M-CONet\cite{wang2023openoccupancy} & L+C & 39.2 & 24.7 & 24.8 & 13.0 & 31.6 & 34.8 & 14.6 & 18.0 & 20.0 & 14.7 & 20.0 & 26.6 & 39.2 & 22.8 & 26.1 & 26.0 & 26.0 & 37.1 \\
Co-Occ\cite{pan_co-occ_2024} & L+C & 41.1 & 27.1 & 28.1 & \underline{16.1} & 34.0 & 37.2 & 17.0 & 21.6 & 20.8 & 15.9 & 21.9 & 28.7 & 42.3 & 25.4 & 29.1 & 28.6 & 28.2 & 38.0 \\
OccMamba\cite{li_occmamba_2025} & L+C & 42.3 & 29.9 & 30.2 & 15.5 & 37.2 & 40.3 & 20.1 & 25.3 & 26.5 & 19.7 & 23.1 & 31.3 & 43.9 & \underline{25.7} & 30.7 & 30.1 & 34.2 & 43.8 \\
SDGOcc\cite{SDGOccDuan_2025_CVPR} & L+C & - & 31.7 & 32.2 & 15.6 & 39.7 & 42.4 & \textbf{22.2} & \textbf{26.9} & 29.7 & 22.7 & \underline{24.2} & 33.2 & 45.9 & 25.1 & 31.8 & \underline{33.5} & \underline{38.9} & 44.5 \\
DAOcc\cite{daocc} & L+C & \underline{42.8} & \underline{32.1} & \underline{32.7} & 15.8 & \underline{40.8} & \underline{42.9} & 21.7 & 26.3 & \underline{30.1} & \textbf{23.4} & 23.7 & \underline{33.5} & \underline{47.2} & 25.4 & \underline{31.9} & \textbf{33.8} & 38.6 & \underline{45.2} \\
\textbf{Gau-Occ (Ours)} & L+C&\textbf{44.3}&\textbf{32.7}&\textbf{33.1} &\textbf{16.5} &\textbf{41.1}&\textbf{43.9} &\underline{21.9}&\underline{26.6}& \textbf{31.1} &\underline{23.1} &\textbf{24.5} &\textbf{34.0} &\textbf{49.0} &\textbf{26.3} &\textbf{32.9} &33.4 &\textbf{39.3 }&\textbf{45.7}\\
\bottomrule
\end{tabular}%
}
\label{tab:comparison_nuscenes}
\end{table*}


\definecolor{others}{cmyk}{0,0,0,1}  
\definecolor{barrier}{cmyk}{0.0,0.53,0.80,0.0}            
\definecolor{bicycle}{cmyk}{0.0,0.25,0.20,0.0}            
\definecolor{bus}{cmyk}{0.0,0.0,1.0,0.0}                  
\definecolor{car}{cmyk}{1.0,0.39,0.0,0.04}                
\definecolor{constveh}{cmyk}{1.0,0.0,0.0,0.0}             
\definecolor{motorcycle}{cmyk}{0.0,0.50,1.0,0.0}          
\definecolor{pedestrian}{cmyk}{0.0,1.0,1.0,0.0}           
\definecolor{trafficcone}{cmyk}{0.0,0.06,0.41,0.0}        
\definecolor{trailer}{cmyk}{0.0,0.56,1.0,0.47}            
\definecolor{truck}{cmyk}{0.33,0.87,0.0,0.06}             
\definecolor{drivesurf}{cmyk}{0.0,1.0,0.0,0.0}            
\definecolor{otherflat}{cmyk}{0.0,0.01,0.01,0.45}         
\definecolor{sidewalk}{cmyk}{0.0,1.0,0.0,0.71}            
\definecolor{terrain}{cmyk}{0.38,0.0,0.67,0.06}           
\definecolor{manmade}{cmyk}{0.08,0.08,0.0,0.02}           
\definecolor{vegetation}{cmyk}{1.0,0.0,1.0,0.31}          

\begin{table*}[t]
\caption{Quantitative comparison on \textbf{Occ3D-nuScenes} validation set. The best results are in \textbf{bold}, second best are \underline{underlined}.
}
\centering
\resizebox{\textwidth}{!}{%
\begin{tabular}{c|c|c|ccccccccccccccccc}
\toprule
\multirow{3}{*}[-0.5em]{Method} & \multirow{3}{*}[-0.5em]{mIoU$\uparrow$} & \multirow{3}{*}[-0.5em]{Modality} &
\makebox[0.8cm]{\rotatebox{90}{others}} & \makebox[0.8cm]{\rotatebox{90}{barrier}} & \makebox[0.8cm]{\rotatebox{90}{bicycle}} & \makebox[0.8cm]{\rotatebox{90}{bus}} &
\makebox[0.8cm]{\rotatebox{90}{car}} & \makebox[0.8cm]{\rotatebox{90}{const. veh.}} & \makebox[0.8cm]{\rotatebox{90}{motorcycle}} & \makebox[0.8cm]{\rotatebox{90}{pedestrian}} &
\makebox[0.8cm]{\rotatebox{90}{traffic cone}} & \makebox[0.8cm]{\rotatebox{90}{trailer}} & \makebox[0.8cm]{\rotatebox{90}{truck}} &
\makebox[0.8cm]{\rotatebox{90}{drive. surf.}} & \makebox[0.8cm]{\rotatebox{90}{other flat}} & \makebox[0.8cm]{\rotatebox{90}{sidewalk}} &
\makebox[0.8cm]{\rotatebox{90}{terrain}} & \makebox[0.8cm]{\rotatebox{90}{manmade}} & \makebox[0.8cm]{\rotatebox{90}{vegetation}} \\
& & &
\makebox[0.8cm]{\colorbox{others}{\rule{0pt}{4pt}\rule{4pt}{0pt}}} &
\makebox[0.8cm]{\colorbox{barrier}{\rule{0pt}{4pt}\rule{4pt}{0pt}}} &
\makebox[0.8cm]{\colorbox{bicycle}{\rule{0pt}{4pt}\rule{4pt}{0pt}}} &
\makebox[0.8cm]{\colorbox{bus}{\rule{0pt}{4pt}\rule{4pt}{0pt}}} &
\makebox[0.8cm]{\colorbox{car}{\rule{0pt}{4pt}\rule{4pt}{0pt}}} &
\makebox[0.8cm]{\colorbox{constveh}{\rule{0pt}{4pt}\rule{4pt}{0pt}}} &
\makebox[0.8cm]{\colorbox{motorcycle}{\rule{0pt}{4pt}\rule{4pt}{0pt}}} &
\makebox[0.8cm]{\colorbox{pedestrian}{\rule{0pt}{4pt}\rule{4pt}{0pt}}} &
\makebox[0.8cm]{\colorbox{trafficcone}{\rule{0pt}{4pt}\rule{4pt}{0pt}}} &
\makebox[0.8cm]{\colorbox{trailer}{\rule{0pt}{4pt}\rule{4pt}{0pt}}} &
\makebox[0.8cm]{\colorbox{truck}{\rule{0pt}{4pt}\rule{4pt}{0pt}}} &
\makebox[0.8cm]{\colorbox{drivesurf}{\rule{0pt}{4pt}\rule{4pt}{0pt}}} &
\makebox[0.8cm]{\colorbox{otherflat}{\rule{0pt}{4pt}\rule{4pt}{0pt}}} &
\makebox[0.8cm]{\colorbox{sidewalk}{\rule{0pt}{4pt}\rule{4pt}{0pt}}} &
\makebox[0.8cm]{\colorbox{terrain}{\rule{0pt}{4pt}\rule{4pt}{0pt}}} &
\makebox[0.8cm]{\colorbox{manmade}{\rule{0pt}{4pt}\rule{4pt}{0pt}}} &
\makebox[0.8cm]{\colorbox{vegetation}{\rule{0pt}{4pt}\rule{4pt}{0pt}}} \\
& & & & & & & & & & & & & & & & & & & \\
\midrule
BEVDetOcc\cite{huang2021bevdet} & 42.0 & C   & 12.2 & 49.6 & 25.1 & 52.0 & 54.5 & 27.9 & 28.0 & 28.9 & 27.2 & 36.4 & 42.2 & 82.3 & 43.3 & 54.6 & 57.9 & 48.6 & 43.6 \\
PanoOcc\cite{wang2024panoocc}    & 42.1 & C   & 11.7 & 50.5 & 29.4 & 49.4 & 55.5 & 24.3 & 33.3 & 30.6 & 31.0 & 34.4 & 42.6 & 83.3 & 44.2 & 54.4 & 56.0 & 45.9 & 40.4 \\
FB-OCC\cite{li2023fbocc}   & 48.9 & C   & 14.3 & 57.0 & 38.3 & 57.7 & 62.1 & 34.4 & 39.4 & 38.8 & 39.4 & 42.9 & 50.0 & 86.0 & 50.2 & \underline{60.1} & \underline{62.5} & 52.4 & 45.7 \\
FlashOcc\cite{yu2023flashocc} & 43.5 & C   & 13.4 & 51.8 & 32.1 & 54.1 & 58.3 & 29.9 & 32.8 & 31.9 & 29.9 & 38.7 & 46.2 & 85.0 & 47.4 & 56.2 & 59.6 & 50.0 & 43.7 \\
COTR  \cite{ma2024cotr}    & 46.2 & C   & 14.9 & 53.3 & 35.2 & 50.8 & 57.4 & 34.1 & 33.5 & 31.9 & 33.1 & 39.1 & 45.0 & 84.5 & 48.3 & 57.6 & 61.1 & 51.6 & 46.7 \\
STCOcc \cite{liao2025stcocc}  & 45.0 & C   & 15.2 & 52.3 & 32.2 & 50.5 & 56.3 & 31.7 & 34.3 & 31.3 & 33.8 & 39.3 & 44.9 & 84.3 & 47.1 & 57.1 & 61.0 & 52.4 & 46.3 \\
ALOCC \cite{chen2025alocc}  & 50.6 & C   & \textbf{16.9} & 55.0 & 37.5 & 56.9 & 62.9 & 32.4 & 39.0 & 34.8 & 34.8 & 41.0 & 50.1 & \underline{87.0} & \underline{53.7} & \textbf{62.7} & \textbf{65.2} & 57.0 & 50.9 \\
RadOcc \cite{zhang2024radocc}  & 49.4 & C   & \underline{15.6} & 55.9 & 35.0 & 52.0 & 62.4 & 33.1 & 38.1 & 32.6 & 33.2 & 39.5 & 48.0 & \textbf{87.2} & 49.8 & 57.6 & \underline{62.5} & 53.7 & 47.7 \\
\midrule
OccFusion \cite{zhang2024occfusion}& 48.7 & C+L+R & 14.2 & 51.8 & 33.3 & 54.0 & 57.7 & 34.0 & 43.0 & 48.4 & 35.5 & 41.2 & 48.6 & 83.4 & \textbf{57.1} & 60.0 & \underline{62.5} & 61.3 & 44.9 \\
SDGOCC\cite{SDGOccDuan_2025_CVPR}& 51.7 & C+L & 13.2 & 57.8 & 34.8 & 60.3 & 64.3 & 36.2 & 39.4 & 52.4 & 35.8 & \underline{50.9} & 53.7 & 84.7 & 47.5 & 58.0 & 61.0 & \underline{70.7} & 67.7 \\
DAOcc\cite{daocc} & \underline{54.3} & C+L & 13.0 & \underline{60.7} & \underline{39.8} & \underline{64.0} & \underline{66.5} & \underline{36.3} & \underline{49.0} & \textbf{60.1} & \textbf{44.3} & 50.7 & \textbf{55.9} & 82.9 & 44.6 & 56.8 & 60.6 & 70.1 & \textbf{68.3} \\
\textbf{Gau-Occ (Ours)} & \textbf{55.1} & C+L & 14.8 & \textbf{61.2} & \textbf{40.7} & \textbf{65.3} & \textbf{68.1} & \textbf{37.5} & \textbf{49.6} & \underline{58.4} & \underline{43.7} & \textbf{51.3} & \underline{55.4} & 84.5 & 48.3 & 58.5 & 60.3 & \textbf{71.5} & \underline{68.0} \\
\bottomrule
\end{tabular}%
}
\label{tab:occ3d_nuscenes_results}
\end{table*}

\begin{figure*}[t]
\centering
\includegraphics[width=0.92\textwidth]{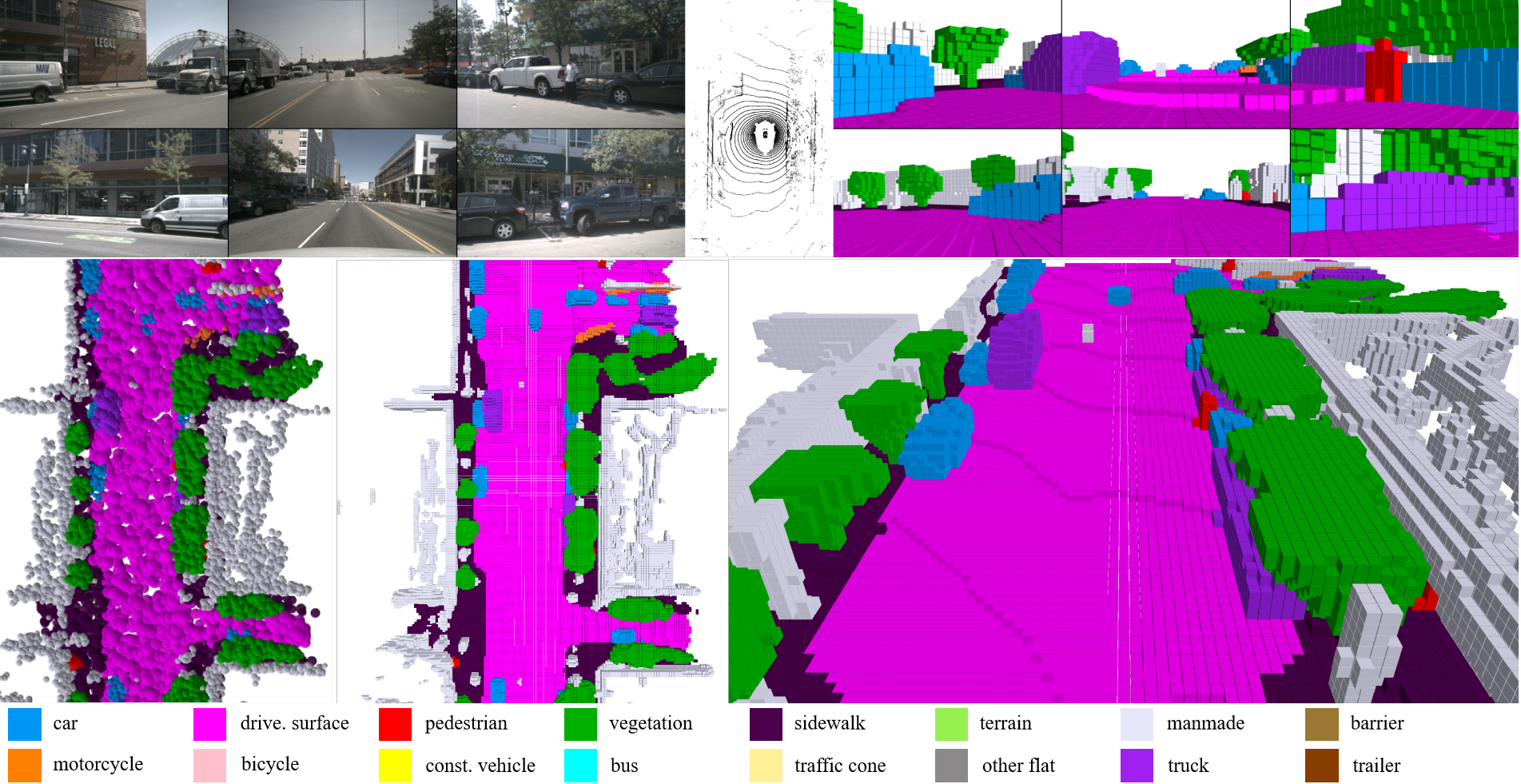} 
\caption{\textbf{Qualitative results on the SurroundOcc-nuScenes validation set.}
Top: multi-view images (left), LiDAR input (center), and predicted image-view occupancy (right). Bottom: predicted 3D Gaussians, BEV occupancy, and front-view occupancy.
}
\label{vis_nuscenes}
\end{figure*}

\begin{figure*}[t]
\centering
\includegraphics[width=0.97\textwidth]{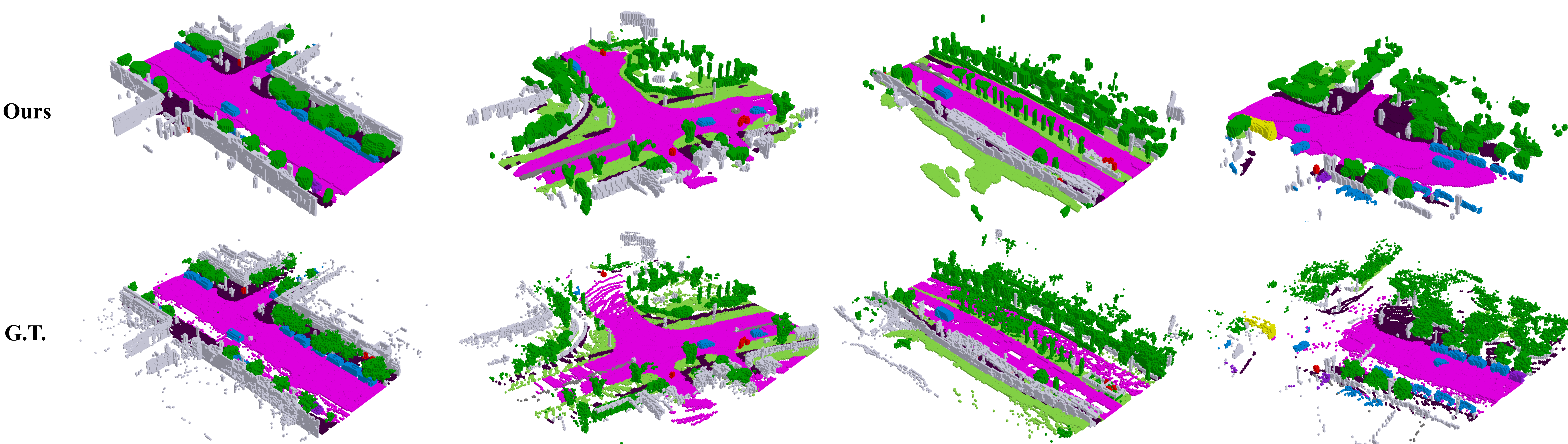} 
\caption{\textbf{Qualitative results on the Occ3D-nuScenes validation set.}
Top: predicted occupancy. Bottom: ground-truth.
}
\label{vis_nuscenes_occ3d}
\end{figure*}


\begin{figure*}[h]
\centering
\includegraphics[width=0.90\textwidth]{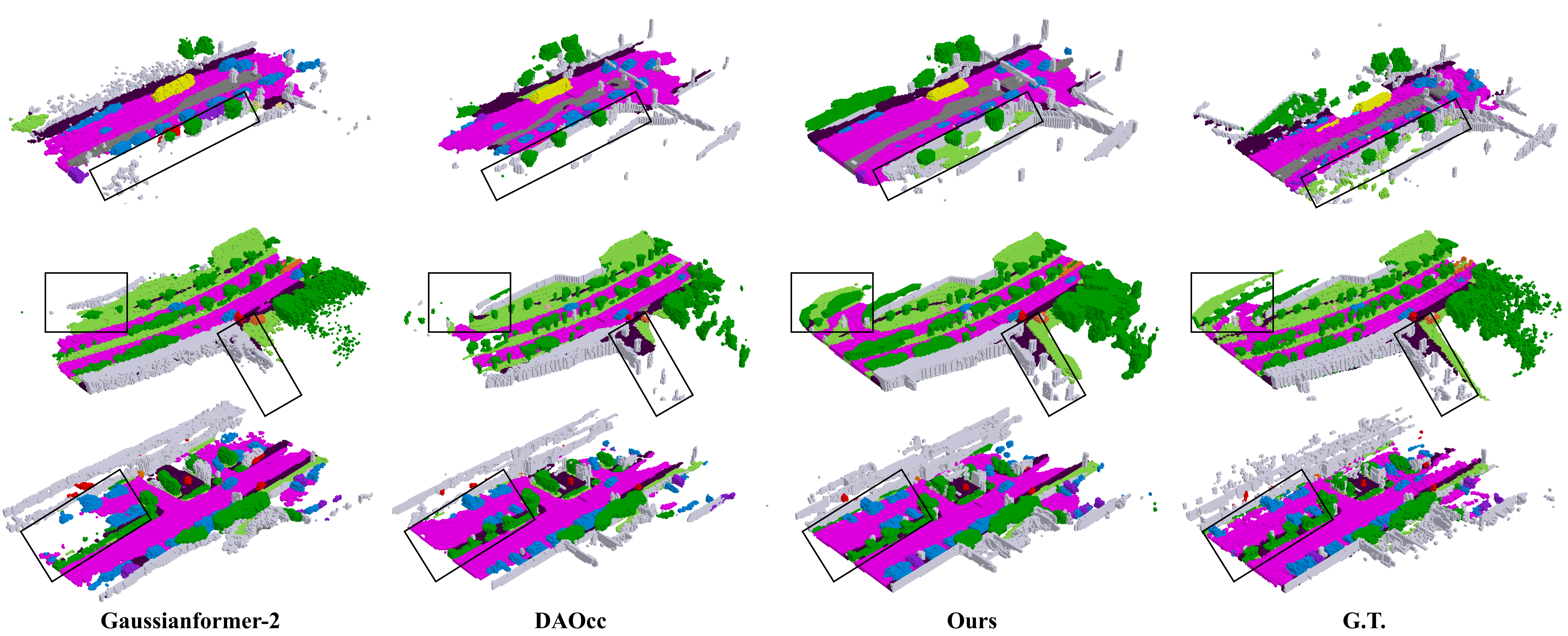}
\caption{Qualitative comparison between Gaussianformer-2~\cite{huang_gaussianformer-2_2025}, DAOcc~\cite{daocc} and Gau-Occ on the SurroundOcc-nuScenes validation set.
}
\label{compare_nuscenes}
\end{figure*}



\subsection{Datasets and Metrics}
We evaluate \textbf{Gau-Occ} on three widely adopted benchmarks: \textit{SurroundOcc-nuScenes}~\cite{caesar2020nuscenes, wei_surroundocc_2023}, \textit{Occ3D-nuScenes}~\cite{tian2023occ3d}, and \textit{KITTI-360}~\cite{liao2022kitti}. 

\subsection{Quantitative Results}
\noindent\textbf{On SurroundOcc-nuScenes.}
Results on the validation split are reported in Tab.~\ref{tab:comparison_nuscenes}, where \textit{L} and \textit{C} denote LiDAR and camera modalities respectively.
Across modalities, LiDAR-only approaches generally outperform camera-only methods due to stronger geometric cues, and multi-modal systems further improve performance. Gau-Occ establishes a new state-of-the-art, surpassing the previous best multi-modal method (DAOcc~\cite{daocc}) by significant margins of \textbf{+1.5 IoU} and \textbf{+0.6 mIoU}.
While DAOcc benefits from detection-level supervision, the proposed Gau-Occ attains superior accuracy without additional priors, highlighting the advantage of geometry-complete Gaussian anchors and structure-aware fusion. 

\noindent\textbf{On Occ3D-nuScenes.}
Under the Occ3D protocol~\cite{tian2023occ3d}, we compare Gau-Occ with strong camera-based systems 
and multi-modal approaches. 
Tab.~\ref{tab:occ3d_nuscenes_results} summarizes the results, where \textit{R} denotes radar.
All methods are evaluated within regions defined by visible mask.
\textbf{Gau-Occ} achieves a new state of the art with \textbf{55.1 mIoU}, surpassing DAOcc by \textbf{+0.8}, SDGOcc by \textbf{+3.4}, and even outperforming radar-augmented OccFusion by \textbf{+6.4}.
This shows that the LiDAR Completion Diffuser (LCD) provides a global geometric prior that enhances not only distant or occluded areas but also visible regions critical for perception, enabling a more complete understanding of scene geometry.
Gau-Occ also achieves clear gains on safety-critical classes such as \textit{bus}, \textit{car}, \textit{bicycle}, and \textit{motorcycle}, benefiting from precise Geo-VLAD resampling and geometry-aware FiLM modulation that align multi-view image evidence with LiDAR-anchored Gaussians and aggregate cues robustly across scale and motion.

\noindent\textbf{On KITTI-360.}
On KITTI-360, we compare Gau-Occ with LiDAR-only methods
and image-only methods. 
Multi-modal baselines are scarce on this dataset.
Owing to page constraints, the complete metric table is deferred to the \textbf{supplementary material}. As shown, Gau-Occ outperforms the strongest LiDAR-only baseline, L2COcc~\cite{wang2025l2cocc}, by \textbf{+1.3 IoU} and \textbf{+0.6 mIoU}. Under this challenging single-camera setting, our method shows notable improvements on moving vehicles (\textit{car}, \textit{truck}) and large structures (\textit{road}, \textit{building}), demonstrating its capability for reliable scene reconstruction from limited visual coverage. These results directly validate the efficacy of our model designs.

\subsection{Qualitative Comparison}
We visualize qualitative results on nuScenes in Fig.~\ref{vis_nuscenes} (SurroundOcc-nuScenes) and Fig.~\ref{vis_nuscenes_occ3d} (Occ3D-nuScenes); additional KITTI-360 visualizations are provided in the \textbf{supplementary material}.
On \emph{SurroundOcc-nuScenes}, Gau-Occ reconstructs fine-scale structures (e.g., pedestrians) while preserving global layout. The LCD module complements sparse LiDAR by generating plausible geometry in occluded and distant regions, leading to more complete scene occupancy. On \emph{Occ3D-nuScenes}, despite denser semantic-occupancy targets, Gau-Occ maintains high structural and semantic consistency. It completes weakly indicated or partially labeled regions and recovers large-scale spatial continuity. 
On \emph{KITTI-360}, under challenging single-camera + LiDAR setting, Gau-Occ maps both large layouts and small instances accurately, demonstrating robustness to sparse viewpoints and effective use of LiDAR geometry.
Fig.~\ref{compare_nuscenes} further provides comparisons with state-of-the-art counterparts, \textit{i.e.} GaussianFormer-2~\cite{huang_gaussianformer-2_2025} and DAOcc~\cite{daocc}, especially at long range and peripheral regions. 
For example, in \emph{Case~1}, Gau-Occ is the only method that reconstructs lower building outlines and terrain surfaces cleanly; 
in \emph{Case~2}, it recovers distant roads and buildings with minimal fragmentation; 
in \emph{Case~3}, it resolves complex road topology without introducing holes in ground or object regions.
These observations support the effectiveness of Gau-Occ’s geometry-complete representation and its robust multi-modal aggregation pipeline.

\begin{table}[t]
\caption{Ablation on the SurroundOcc-nuScenes validation set.}
\label{tab:ablation}
\centering
\setlength{\tabcolsep}{5pt}
\renewcommand{\arraystretch}{1.12}

\begin{subtable}{\linewidth}
\caption{Impact of point cloud source and Gaussian initialization. 
$\mathcal{P}$: raw LiDAR, PD($\mathcal{P}$): LiDPM~\cite{martyniuk2025lidpm} completion. 
$\mathcal{P}'$: our completion. DS: density-based selection, RS: random coverage sampling.}
\centering
{\small 
\begin{tabular}{c|c|c|cc}
\toprule
\# &\quad Point Cloud  \quad\quad & \quad Gaussian Init. \quad\quad & IoU$\uparrow$ & mIoU$\uparrow$ \\
\midrule
1 & $\mathcal{P}$        & DS + RS & 41.5 & 29.6 \\
2 & PD$(\mathcal{P})$    & DS + RS & 43.1 & 31.9 \\
3 & $\mathcal{P}'$       & RS      & 43.9 & 32.4 \\
4 & $\mathcal{P}'$       & DS + RS & \textbf{44.3} & \textbf{32.7} \\
\bottomrule
\end{tabular}
}
\label{tab:ablation-a}
\end{subtable}


\begin{subtable}{\linewidth}
\caption{Ablation of \textbf{GAF} components on nuScenes.}
\centering
{\small
\begin{tabular}{c|c|c|c|cc}
\toprule
\# & \quad GAF \quad \quad &\quad  GGS \quad \quad & \quad GVR \quad \quad & IoU$\uparrow$ & mIoU$\uparrow$ \\
\midrule
1 &      &       &          & 35.2 & 24.9 \\
2 & \checkmark &       & \checkmark & 40.6 & 31.2 \\
3 & \checkmark & \checkmark &          & 43.9 & 32.4 \\
4 & \checkmark & \checkmark & \checkmark & \textbf{44.3} & \textbf{32.7} \\
\bottomrule
\end{tabular}
}
\label{tab:ablation-b}
\end{subtable}

\end{table}

\subsection{Ablation Study}
\begin{figure}[t]
\centering
\includegraphics[width=0.95\linewidth]{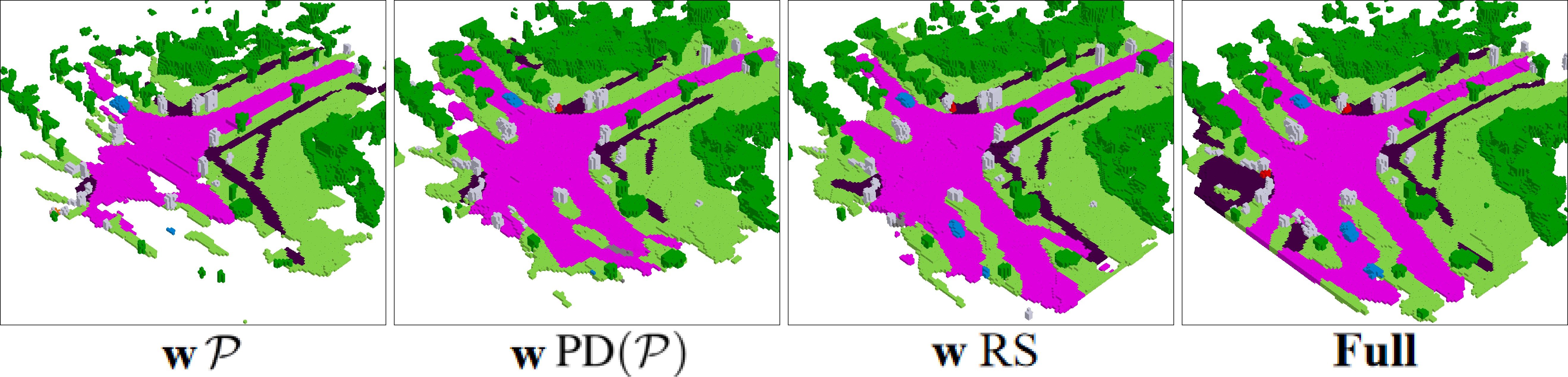} 
\caption{Visualization of ablations on Gaussian initialization.}
\label{abla2:a}
\end{figure}

\noindent\textbf{On Point Cloud Completion and Gaussian Initialization.} As shown in Tab.~\ref{tab:ablation-a} and Fig.~\ref{abla2:a}, replacing the completed point cloud $\mathcal{P}'$ with the raw input $\mathcal{P}$ leads to notable performance drops in both IoU and mIoU. This confirms that LCD module significantly enhances scene coverage for distant regions and occluded road surfaces. Compared to diffusion-based alternatives such as LiDPM~\cite{martyniuk2025lidpm} (omitted for brevity), our lightweight pre-trained module provides superior geometric priors. Furthermore, the hybrid initialization strategy combining DS (density-based selection) and RS (random sampling) consistently outperforms the use of vanilla RS alone. This approach balances structural concentration with broad scene coverage, enabling better reconstruction of far-range and easily overlooked object classes.

\textbf{On Gaussian Anchor Fusion (GAF).}
We conduct a comprehensive ablation study on the GAF module, focusing on two core components governing cross-modal fusion:
(1) \textbf{GGS} (Geometry-Guided Sampling), which conditions 2D sampling offsets on LiDAR features $\mathbf{f}_{pc}$ for spatially-aware feature retrieval;
(2) \textbf{GVR} (Geo-VLAD Resampling), a codebook-based residual aggregator that compresses the sampled tokens $\mathbf{X}_i\!\in\!\mathbb{R}^{N\times d}$ into $\mathbf{Z}_i\!\in\!\mathbb{R}^{M\times d}$ (here $M{=}32$, $N{=}216$) via LiDAR-conditioned soft assignments.

\begin{figure}[t]
\centering
\includegraphics[width=0.95\linewidth]{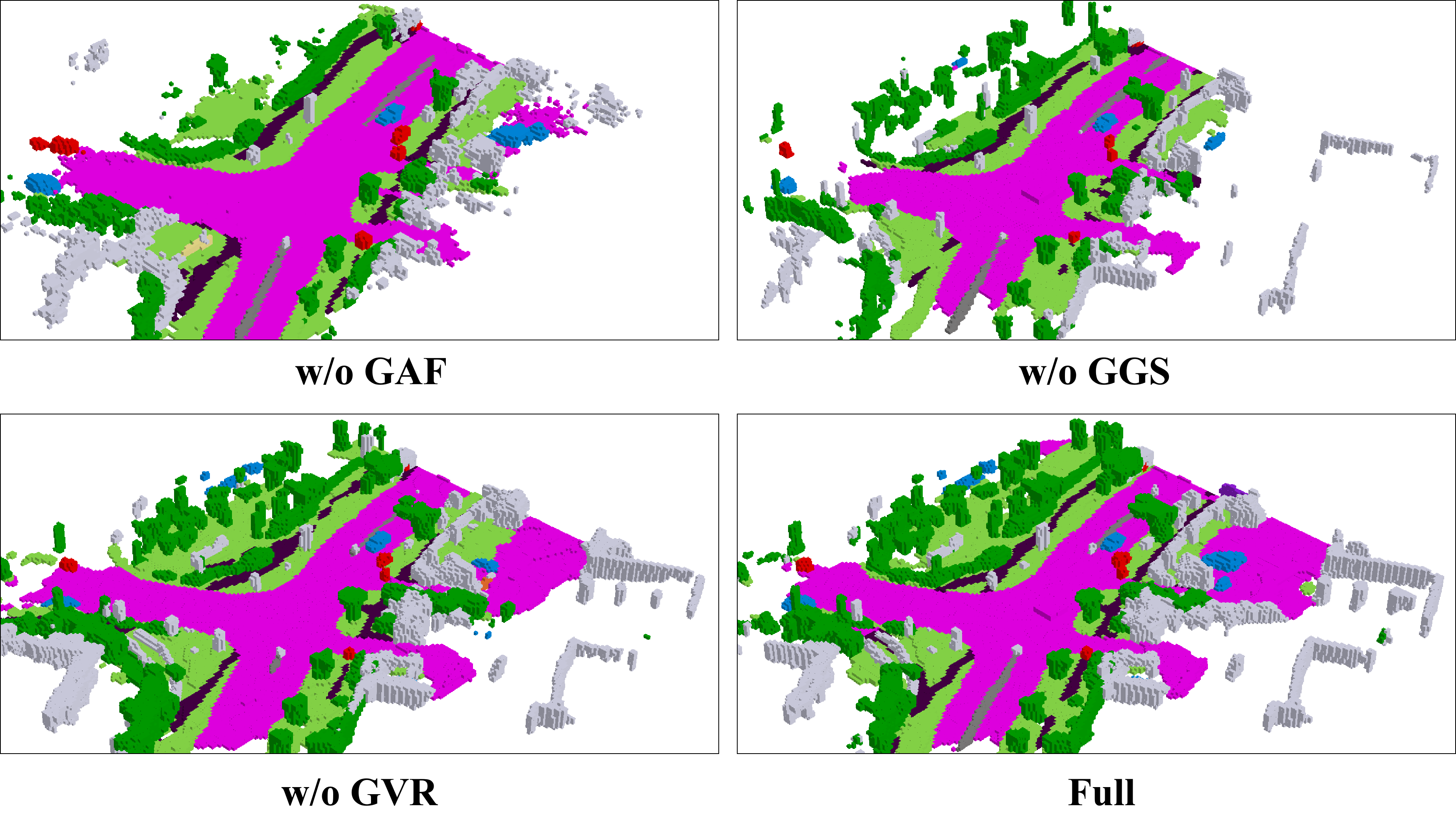}
\caption{Visualization of ablations on \textbf{GAF} components.}
\label{abla1:a}
\end{figure}

As summarized in Tab.~\ref{tab:ablation-b}, the absence of GAF (Row 1) leads to a significant performance drop, confirming that image-only feature extraction combined with point clouds used only for initialization substantially hinders representation capability.
Replacing GGS with geometry-agnostic sampling (Row 2) degrades long-range feature association, underscoring the importance of LiDAR-conditioned offsets in maintaining spatial and semantic consistency during fusion.
Removing \textbf{GVR} (Row 3) and directly feeding the original, unaggregated tokens $\mathbf{X}_i$ to cross-attention leads to markedly higher latency and memory usage. This causes a slight accuracy drop, attributed to token redundancy.
The full GAF configuration (Row 4) achieves optimal results, validating the necessity of both geometry-guided sampling and refinement in building a robust multi-modal representation.
Qualitative ablation results are shown in Fig.~\ref{abla1:a}.

\section{Conclusion}
This paper introduces Gau-Occ, a multi-modal 3D semantic occupancy framework based on semantic 3D Gaussians. It leverages the LCD module for geometry completion from sparse LiDAR and the GAF module for efficient, geometry-guided image feature aggregation. Evaluated on multiple benchmarks, Gau-Occ achieves state-of-the-art results with high computational efficiency, proving its effectiveness.

\section*{Acknowledgment}

This work is partially supported by  the New Generation Artificial Intelligence-National Science and Technology Major Project (2025ZD0124000), Xiaomi Young Talents Program, the Research Program of State Key Laboratory of Virtual Reality Technology and Systems, and the Fundamental Research Funds for the Central Universities.
{
    \small
    \bibliographystyle{ieeenat_fullname}
    \bibliography{main}
}

\clearpage
\setcounter{page}{1}
\maketitlesupplementary

\section{Datasets and Metrics}

\noindent\textbf{nuScenes and SurroundOcc-nuScenes.} nuScenes is a large-scale autonomous-driving dataset collected in Boston and Singapore that contains over $1{,}000$ urban scenes. Each scene lasts roughly $20$ seconds and is captured with six surround-view cameras and one LiDAR sensor; keyframes are annotated at $2$ Hz. Following SurroundOcc~\cite{wei_surroundocc_2023}, we adopt dense semantic-occupancy annotations that voxelize the region $[-50,50]~\text{m}\times[-50,50] ~\text{m}\times[-5,3]~\text{m}$ with $0.5$ m voxel resolution, assigning one of 18 classes (16 semantic categories, plus \texttt{empty} and \texttt{unknown}) to every voxel. We use the official nuScenes split (700/150/150 train/val/test) and follow the standard sensor configuration and annotation protocol. Accordingly, for each keyframe our input consists of the synchronized six-camera images and the LiDAR sweep, and the target is the voxelized 3D occupancy ground-truth.

\noindent\textbf{Occ3D-nuScenes.}
To further evaluate our method under a distinct semantic-occupancy protocol derived from nuScenes, we also consider Occ3D-nuScenes. Following the official Occ3D devkit~\cite{tian2023occ3d}, we construct voxel volumes over $[-40,40]\,\text{m} \times [-40,40]\,\text{m} \times [-1,5.4]\,\text{m}$ at a $0.4$\,\text{m} resolution, with 17 semantic categories (16 base classes plus a \texttt{General Object} class). We use the standard 600/150/150 train/val/test split, totaling 40{,}000 annotated keyframes. Similar to SurroundOcc-nuScenes, each keyframe provides six surround-view camera images and a single LiDAR scan. However, the differing voxel range, resolution, and label space make Occ3D-nuScenes a complementary benchmark for assessing robustness to changes in occupancy definitions and grid configurations.

\noindent\textbf{KITTI-360.}
KITTI-360 offers over 320k multi-view images and 100k LiDAR sweeps from long urban drives. We adopt the dense semantic occupancy annotations released by SSCBench-KITTI-360~\cite{li2024sscbench}, which provide ground-truth semantic occupancy for 12{,}865 keyframes across nine sequences with the standard 7/1/1 train/validation/test split. The voxel grid covers $[0, 51.2]\,\text{m} \times [-25.6, 25.6]\,\text{m} \times [-2, 4.4]\,\text{m}$ at a $0.2$\,\text{m} resolution, with each voxel labeled as one of 19 categories (18 semantic classes plus \texttt{empty}). Following common practice, we use only the left-front perspective camera (image\_00 subset) of each keyframe together with the corresponding raw LiDAR point cloud as model input.

We adopt standard evaluation metrics for semantic occupancy prediction tasks. 
Following common practice, we use the \textbf{Intersection over Union (IoU)} of all occupied voxels to evaluate the geometry reconstruction performance of the model, and the \textbf{mean Intersection over Union (mIoU)} of all semantic classes to evaluate its semantic perception ability.
The IoU and mIoU are computed as follows:
\begin{equation}
\label{eq:iou_def}
\mathrm{IoU} = \frac{TP_{c_0}}{TP_{c_0} + FP_{c_0} + FN_{c_0}},
\end{equation}
\begin{equation}
\label{eq:miou_def}
\mathrm{mIoU} = \frac{1}{|C|} \sum_{i \in C} \frac{TP_i}{TP_i + FP_i + FN_i},
\end{equation}
where $c_0$ denotes the nonempty (occupied) class; $TP_i$, $FP_i$, and $FN_i$ are the number of true positive, false positive, and false negative predictions for class $i$, $C$ is the set of semantic classes.
These metrics jointly provide a comprehensive evaluation of both geometric reconstruction quality and semantic occupancy prediction accuracy.

\section{Experimental Setup}

\paragraph{LCD pre-training.}
The \emph{LiDAR Completion Diffuser (LCD)} is pre-trained on dense targets built by ego-motion alignment and accumulation of $K{=}20$ consecutive sweeps.
We train LCD for $20$ epochs on the respective training split before joint optimization with the proposed Gau-Occ framework.
The forward process follows DDPM with $T{=}1000$ steps and a linear schedule $\{\beta_t\}_{t=1}^{T}$ (default $\beta_0{=}3.0{\times}10^{-5}$, $\beta_T{=}7.0{\times}10^{-3}$), $\alpha_t{=}1{-}\beta_t$, $\bar\alpha_t{=}\prod_{i=1}^{t}\alpha_i$.
During training, we use DPM-Solver sampling with $50$ denoising steps.
All hyper-parameters above are held fixed across datasets unless stated.

\paragraph{Semantic Gaussians.}
We instantiate a dataset-specific number of semantic Gaussians: $N_G{=}25{,}600$ for nuScenes and $N_G{=}40{,}000$ for KITTI-360. 
Hybrid initialization selects centers from the completed cloud $\mathcal{P}'$ via density-based selection and random coverage: the default split is $N_d{:}N_r{=}70\%{:}30\%$.
Each new Gaussian has an axis-aligned initial scale $\mathbf{s}_i\!\sim\!\mathcal{U}([0.20,\,1.00])$ per axis.
Local splatting uses a neighborhood radius
$R_{\mathrm{geo}}{=}k\,\overline{s}_i$ with $\overline{s}_i{=}\tfrac{1}{3}(s_x{+}s_y{+}s_z)$ and default $k{=}1.5$.

\paragraph{LiDAR voxel features.}
We voxelize the completed cloud \(\mathcal{P}'\) into a sparse 3D grid (bounds and voxel size as in the main paper) and keep at most \(10\) points per voxel~\cite{yan2018second}. Per-voxel features \(\mathbf{F}_v\) are obtained by averaging point embeddings \(\psi(p)\) to \(\mathbf{f}^0_v\) and feeding a sparse 3D CNN encoder that outputs \(d_{\!pc}\)-dimensional descriptors where \(d_{\!pc}{=}128\). For a Gaussian \(G_i\) centered at \(\boldsymbol{\mu}_i\) with scale \(\mathbf{s}_i{=}(s_x,s_y,s_z)\), we aggregate neighboring voxels within an adaptive radius \(R_{\mathrm{geo}}{=}k\,(s_x{+}s_y{+}s_z)/3\) where \(k{=}1.5\),  using an exponential kernel \(w_v{=}\exp(-\gamma\|\mathbf{p}_v{-}\boldsymbol{\mu}_i\|_2)\) (default \(\gamma{=}3.0\)), yielding the geometry descriptor \(\mathbf{f}_{\mathrm{pc},i}\in\mathbb{R}^{d_{\!pc}}\).

\paragraph{Image backbone and pyramid.}
Unless otherwise noted, we use ResNet-50 with a 4-level FPN (\(L{=}4\)) at strides \(s_l\!\in\!\{4,8,16,32\}\). Each level has channel width \(d{=}128\). The number of camera views is dataset-specific: \(V{=}6\) for nuScenes and \(V{=}1\) for KITTI-360 (image\_00). Each anchor predicts \(N_{\text{off}}{=}9\) geometry-guided offsets per level/view; sampling radius are \(R_l\!\in\!\{4,8,16,32\}\) feature pixels . The geometry weight uses \(\sigma_l{=}\kappa R_l\) with default \(\kappa{=}1.0\). 

\paragraph{Geo-VLAD resampler, cross-attention, and update head.}
Sampled tokens \(\mathbf{X}_i\!\in\!\mathbb{R}^{N\times d}\) are compressed by a geometry-aware VLAD-style resampler with \(M\) codewords \(\{\mathbf{C}_m\}_{m=1}^{M}\), where \(M{=}32\). Linear maps follow the shapes in the main text. FiLM modulation predicts per-channel \((\gamma_i,\beta_i)\) from \(\mathbf{f}_{\mathrm{pc},i}\) to rescale/shift the resampled tokens; multi-scale fusion uses learnable non-negative weights \(\{\lambda_l\}_{l=1}^{L}\) (softmax-normalized). The Gaussian update head is a two-layer feed-forward network (FFN) with GELU and hidden size \(128\), regressing \([\widehat{\boldsymbol{\mu}}_i,\widehat{\mathbf{s}}_i,\widehat{\mathbf{r}}_i,\widehat{\mathbf{c}}_i]\); updated Gaussians \(\mathbf{G}_i^{\text{new}}{=}(\boldsymbol{\mu}_i{+}\widehat{\boldsymbol{\mu}}_i,\widehat{\mathbf{s}}_i,\widehat{\mathbf{r}}_i,\widehat{\mathbf{c}}_i)\) are then splatted locally to produce \(O\).

\paragraph{Optimization and implementation details.}
We minimize $\mathcal{L}_{\text{CE}}{+}\mathcal{L}_{\text{Lov}}$ following \cite{huang_tri-perspective_2023}.
AdamW is used with weight decay $0.01$; the learning rate warms up linearly for $500$ iters to $2{\times}10^{-4}$ and then follows cosine decay to $1{\times}10^{-6}$. 
Unless specified, training runs for $20$ epochs on nuScenes and $25$ on KITTI-360 with batch size $8$.
We implement in PyTorch 1.12.1 (Python 3.9, Ubuntu 22.04).
nuScenes experiments are trained/inferred on RTX 4090 (24\,GB); KITTI-360 on A100 (40\,GB).

\section{Model Efficiency}

\begin{table}[t]
\centering
\resizebox{\columnwidth}{!}{%
\begin{tabular}{c|c|c|c|c|c|c}
\toprule
\multirow{2}{*}{\textbf{Method}} & \multirow{2}{*}{\textbf{Modality}}& \textbf{Query} & \textbf{Lat.} & \textbf{Mem.} & \multirow{2}{*}{\textbf{IoU}$\uparrow$} & \multirow{2}{*}{\textbf{mIoU}$\uparrow$} \\
 & & \textbf{Number} & \textbf{(ms)} & \textbf{(GB)} & & \\
\midrule
BEVFormer\cite{li2022bevformer} & C & 200×200 & 310 & 4.5 & 30.5 & 16.8 \\
TPVFormer\cite{huang_tri-perspective_2023} &  C & 200×200×16 & 320 & 5.1 & 30.9 & 17.1 \\
SurroundOcc\cite{wei_surroundocc_2023}  &C  & 200×200×16 & 340 & 5.9 & 31.5 & 16.3 \\
\cmidrule{3-6}
\multirow{2}{*}{GaussianFormer\cite{huang_gaussianformer_2025}} &  C & 25600 & 195 & 4.8 & 28.7 & 16.0 \\
 &  C& 144000 & 372 & 6.2 & 29.8 & 19.1 \\
\cmidrule{3-6}
\multirow{2}{*}{GaussianFormer-2\cite{huang_gaussianformer-2_2025}} & C &  12800 & 323 & \textbf{3.0} & 30.4 & 19.9 \\
 &  C& 25600 & 357 & \textbf{3.0} & 31.0 & 20.3 \\
\midrule
M-CONet\cite{wang2023openoccupancy}  & L+C &100×100×8 & 670 & 7.8 & 39.2 & 24.7 \\
Co-Occ\cite{pan_co-occ_2024} & L+C& 100×100×8 & 595 & 12.1 & 41.1 & 27.1 \\
DAOcc*\cite{daocc} & L+C& 456×456 & 130 & 4.2 & 40.5 & 30.3 \\
DAOcc\cite{daocc} & L+C& 720×720 & 291 & 8.6 & 42.8 & 32.1 \\
\midrule
\multirow{2}{*}{\textbf{Ours}}& L+C & 12800 & \textbf{124} & 3.3 & 42.4 & 31.5 \\
& L+C & 25600 & 230 & 5.4 & \textbf{44.3} & \textbf{32.7} \\
\bottomrule
\end{tabular}%
}
\caption{
Comparison of inference efficiency on the nuScenes validation set. All results are measured with a batch size of 1 on a single NVIDIA RTX 4090 GPU.
}
\label{latency_memory}
\end{table}

\noindent
Tab.~\ref{latency_memory} compares the \textit{latency, memory, accuracy} trade-off of different 3D occupancy prediction pipelines.

For the camera-only baselines at the top of the table, BEVFormer, TPVFormer, and SurroundOcc all rely on dense BEV or volumetric queries, leading to relatively high computational cost, they run at 310 $\sim$ 340 ms with 4.5 $\sim$ 5.9 GB memory, while only achieving around 31 IoU and 16 $\sim$ 17 mIoU.
Under the 12.8k-query setting, our model attains 124 ms latency and 3.3 GB memory, which is about $2.5\times$ faster and $27\sim44\%$ more memory-efficient than these BEV-based camera-only methods, while delivering much higher IoU and mIoU.
Even the higher-parameter 25.6k-query configuration still runs faster than BEV-based camera models (230 ms vs. 310$\sim$340 ms) with comparable or lower memory (5.4 GB), and sets the best overall accuracy (44.3 IoU and 32.7 mIoU). GaussianFormer and GaussianFormer-2 are camera-only Gaussian-query baselines that process multi-view tokens with several global transformer blocks. Their IoU/mIoU are competitive within the camera-only group but remain below multi-modal entries in Tab.~\ref{latency_memory}.

\definecolor{car}{cmyk}{0.59,0.39,0.0,0.04}               
\definecolor{bicycle}{cmyk}{0.59,0.10,0.0,0.04}            
\definecolor{motorcycle}{cmyk}{0.80,0.60,0.0,0.41}         
\definecolor{truck}{cmyk}{0.56,0.83,0.0,0.29}              
\definecolor{othervehicle}{cmyk}{1.0,1.0,0.0,0.0}          
\definecolor{person}{cmyk}{0.0,0.88,0.88,0.0}              
\definecolor{road}{cmyk}{0.0,1.0,0.0,0.0}                  
\definecolor{parking}{cmyk}{0.0,0.41,0.0,0.0}              
\definecolor{sidewalk}{cmyk}{0.0,1.0,0.0,0.71}             
\definecolor{otherground}{cmyk}{0.0,1.0,0.57,0.31}         
\definecolor{building}{cmyk}{0.0,0.22,1.0,0.0}             
\definecolor{fence}{cmyk}{0.0,0.53,0.80,0.0}               
\definecolor{vegetation}{cmyk}{1.0,0.0,1.0,0.31}           
\definecolor{terrain}{cmyk}{0.38,0.0,0.67,0.06}            
\definecolor{pole}{cmyk}{0.0,0.06,0.41,0.0}                
\definecolor{trafficsign}{cmyk}{0.0,1.0,1.0,0.0}           
\definecolor{otherstructure}{cmyk}{0.0,0.41,1.0,0.0}       
\definecolor{otherobject}{cmyk}{0.80,0.0,0.0,0.0}          

\begin{table*}[t]
\centering
\vspace{-0.3cm}
\caption{Quantitative comparison on the KITTI-360 validation set. The best results are in \textbf{bold}, second best are \underline{underlined}.}
\label{tab:comparison_kitti}
\resizebox{\textwidth}{!}{%
\begin{tabular}{c|c|cc|cccccccccccccccccc}
\toprule
\multirow{3}{*}[-0.5em]{Method} & \multirow{3}{*}[-0.5em]{Modality} & \multirow{3}{*}[-0.5em]{\makebox[1cm]{IoU$\uparrow$}} & \multirow{3}{*}[-0.5em]{\makebox[1cm]{mIoU$\uparrow$}} & \makebox[0.8cm]{\rotatebox{90}{car}} & \makebox[0.8cm]{\rotatebox{90}{bicycle}} & \makebox[0.8cm]{\rotatebox{90}{motorcycle}} & \makebox[0.8cm]{\rotatebox{90}{truck}} & \makebox[0.8cm]{\rotatebox{90}{other-vehicle}} & \makebox[0.8cm]{\rotatebox{90}{person}} & \makebox[0.8cm]{\rotatebox{90}{road}} & \makebox[0.8cm]{\rotatebox{90}{parking}} & \makebox[0.8cm]{\rotatebox{90}{sidewalk}} & \makebox[0.8cm]{\rotatebox{90}{other-ground}} & \makebox[0.8cm]{\rotatebox{90}{building}} & \makebox[0.8cm]{\rotatebox{90}{fence}} & \makebox[0.8cm]{\rotatebox{90}{vegetation}} & \makebox[0.8cm]{\rotatebox{90}{terrain}} & \makebox[0.8cm]{\rotatebox{90}{pole}} & \makebox[0.8cm]{\rotatebox{90}{traffic-sign}} & \makebox[0.8cm]{\rotatebox{90}{other-structure}} & \makebox[0.8cm]{\rotatebox{90}{other-object}} \\
& & & & \makebox[0.8cm]{\colorbox{car}{\rule{0pt}{4pt}\rule{4pt}{0pt}}} & \makebox[0.8cm]{\colorbox{bicycle}{\rule{0pt}{4pt}\rule{4pt}{0pt}}} & \makebox[0.8cm]{\colorbox{motorcycle}{\rule{0pt}{4pt}\rule{4pt}{0pt}}} & \makebox[0.8cm]{\colorbox{truck}{\rule{0pt}{4pt}\rule{4pt}{0pt}}} & \makebox[0.8cm]{\colorbox{othervehicle}{\rule{0pt}{4pt}\rule{4pt}{0pt}}} & \makebox[0.8cm]{\colorbox{person}{\rule{0pt}{4pt}\rule{4pt}{0pt}}} & \makebox[0.8cm]{\colorbox{road}{\rule{0pt}{4pt}\rule{4pt}{0pt}}} & \makebox[0.8cm]{\colorbox{parking}{\rule{0pt}{4pt}\rule{4pt}{0pt}}} & \makebox[0.8cm]{\colorbox{sidewalk}{\rule{0pt}{4pt}\rule{4pt}{0pt}}} & \makebox[0.8cm]{\colorbox{otherground}{\rule{0pt}{4pt}\rule{4pt}{0pt}}} & \makebox[0.8cm]{\colorbox{building}{\rule{0pt}{4pt}\rule{4pt}{0pt}}} & \makebox[0.8cm]{\colorbox{fence}{\rule{0pt}{4pt}\rule{4pt}{0pt}}} & \makebox[0.8cm]{\colorbox{vegetation}{\rule{0pt}{4pt}\rule{4pt}{0pt}}} & \makebox[0.8cm]{\colorbox{terrain}{\rule{0pt}{4pt}\rule{4pt}{0pt}}} & \makebox[0.8cm]{\colorbox{pole}{\rule{0pt}{4pt}\rule{4pt}{0pt}}} & \makebox[0.8cm]{\colorbox{trafficsign}{\rule{0pt}{4pt}\rule{4pt}{0pt}}} & \makebox[0.8cm]{\colorbox{otherstructure}{\rule{0pt}{4pt}\rule{4pt}{0pt}}} & \makebox[0.8cm]{\colorbox{otherobject}{\rule{0pt}{4pt}\rule{4pt}{0pt}}} \\
& & & & & & & & & & & & & & & & & & & & & \\
\midrule
LMSCNet\cite{roldao2020lmscnet} & L & 47.5 & 13.7 & 20.9 & 0.0 & 0.0 & 0.3 & 0.0 & 0.0 & 63.0 & 13.5 & 33.5 & 0.2 & 43.7 & 0.3 & 40.0 & 26.8 & 0.0 & 0.0 & 3.6 & 0.0 \\
SSCNet\cite{song_semantic_2017} & L & 53.6 & 17.0 & 32.0 & 0.0 & 0.2 & 10.3 & 0.6 & 0.1 & 65.7 & 17.3 & 41.2 & 3.2 & 44.4 & 6.8 & 43.7 & 28.9 & 0.8 & 0.8 & 8.6 & 0.7 \\
L2COcc-L\cite{wang2025l2cocc} & L & \underline{57.6} & \underline{25.2} & \underline{40.4} & \textbf{4.1} & 4.2 & \underline{26.2} & \underline{10.3} & \underline{10.0} & \underline{70.0} & \textbf{22.6} & \underline{46.2} & \underline{8.8} & \underline{51.0} & \underline{14.3} & \textbf{47.6} & \underline{31.3} & \underline{24.9} & \underline{21.6} & \underline{14.0} & \underline{6.4} \\
\midrule
MonoScene\cite{cao_monoscene_2022}& C & 37.9 & 12.3 & 19.3 & 0.4 & 0.6 & 8.0 & 2.0 & 0.9 & 48.4 & 11.4 & 28.1 & 3.2 & 32.9 & 3.5 & 26.2 & 16.8 & 6.9 & 5.7 & 4.2 & 3.1 \\
VoxFormer\cite{li_voxformer_2023}& C & 38.8 & 11.9 & 17.8 & 1.2 & 0.9 & 4.6 & 2.1 & 1.6 & 47.0 & 9.7 & 27.2 & 2.8 & 31.2 & 5.0 & 29.0 & 14.7 & 6.5 & 6.9 & 3.8 & 2.4 \\
TPVFormer\cite{huang_tri-perspective_2023} & C & 40.2 & 13.7 & 21.6 & 1.1 & 1.4 & 8.1 & 2.6 & 2.4 & 53.0 & 12.0 & 31.1 & 3.8 & 34.8 & 4.8 & 30.1 & 17.5 & 7.5 & 5.9 & 5.5 & 2.7 \\
OccFormer\cite{zhang_occformer_2023} & C & 40.3 & 14.6 & 22.6 & 0.7 & 0.3 & 9.9 & 3.8 & 2.8 & 54.3 & 13.4 & 31.5 & 3.6 & 36.4 & 4.8 & 31.0 & 19.5 & 7.8 & 8.5 & 7.0 & 4.6 \\
Gaussianformer\cite{huang_gaussianformer_2025} & C & 35.4 & 12.9 & 18.9 & 1.0 & \textbf{4.6} & 18.1 & 7.6 & 3.4 & 45.5 & 10.9 & 25.0 & 5.3 & 28.4 & 5.7 & 29.5 & 8.6 & 3.0 & 2.3 & 9.5 & 5.1 \\
Gaussianformer-2\cite{huang_gaussianformer-2_2025} & C & 38.4 & 13.9 & 21.1 & 2.6 & 4.2 & 12.4 & 5.7 & 1.6 & 54.1 & 11.0 & 32.3 & 3.3 & 32.0 & 5.0 & 28.9 & 17.3 & 3.6 & 5.5 & 5.9 & 3.5 \\
L2COcc-C\cite{wang2025l2cocc}  & C & 48.1 & 20.1 & 29.6 & \underline{3.7} & 4.4 & 14.9 & 8.4 & 7.2 & 63.3 & 17.9 & 40.5 & 5.2 & 42.8 & 8.5 & 39.4 & 24.5 & 16.2 &18.4 & 10.2 & 6.8 \\
\midrule
\textbf{Gau-Occ (Ours)} & L+C & \textbf{58.9} & \textbf{25.8} & \textbf{42.1} & \underline{3.7} & \underline{4.4} & \textbf{27.2} & \textbf{10.8} & \textbf{10.5} & \textbf{70.9} & \underline{22.5} & \textbf{47.1} & \textbf{9.4} & \textbf{51.7} & \textbf{14.6} & \underline{47.0} & \textbf{32.2} & \textbf{25.2} & \textbf{22.7} & \textbf{14.1} & \textbf{7.8} \\
\bottomrule
\end{tabular}%
}
\end{table*}

\begin{figure*}[t]
\centering
\includegraphics[width=0.99\linewidth]{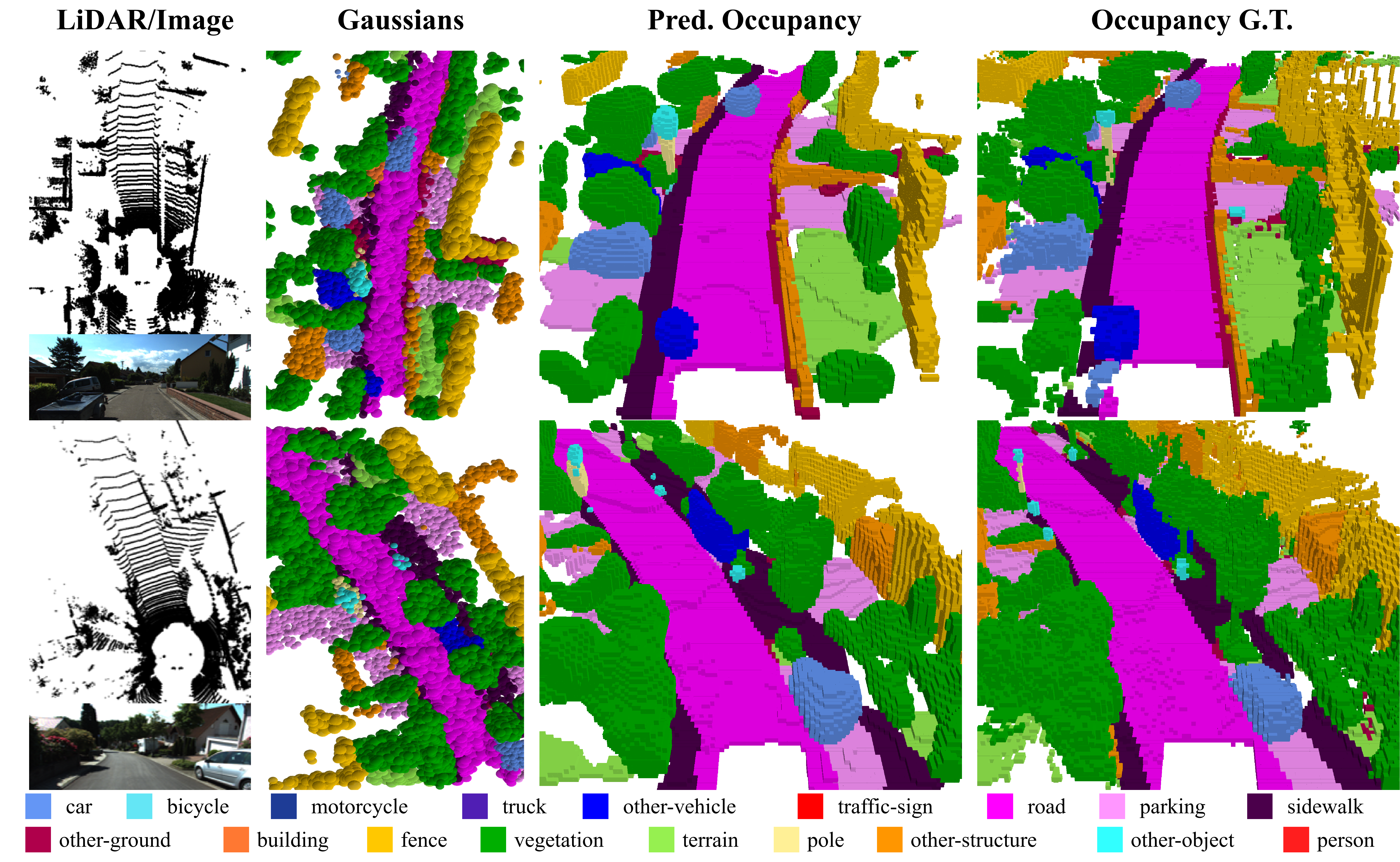} 

\caption{Qualitative results on KITTI-360.
}
\label{vis_kitti360}
\vspace{-0.3cm}
\end{figure*}

\noindent
We next compare with the multi-modal methods M-CONet and Co-Occ.
Both methods fuse LiDAR and camera inputs with dense BEV queries (100 × 100 × 8) and therefore incur substantial latency and memory. M-CONet requires 670 ms and 7.8 GB to reach 39.2/24.7 IoU/mIoU, while Co-Occ takes 595 ms and 12.1 GB for 41.1/27.1.
In contrast, our sparse Gaussian design is substantially more efficient in both time and space while achieving clearly better accuracy.
With only 12.8k Gaussian queries, our model runs at 124 ms and 3.3 GB, which is about $5.4\times$ and $4.8\times$ faster than M-CONet (670 ms) and Co-Occ (595 ms), respectively, while reducing memory consumption by roughly $58\%$ and $73\%$ (from 7.8 GB and 12.1 GB to 3.3 GB), and simultaneously improving IoU/mIoU from 39.2/24.7 and 41.1/27.1 to 42.4/31.5.

\noindent
The more recent method DAOcc adopts efficient per-query operations but relies on a very dense BEV grid with 720 × 720 queries (over $5\times$ more queries than M-CONet/Co-Occ and over $40\times$ more than our 12.8k-Gaussian setting), which leads to a total latency of 291 ms and 8.6 GB memory.
Although the downsampled variant DAOcc* reduces the BEV resolution to 456 × 456, lowering latency and memory to 130 ms and 4.2 GB, it still remains slower and heavier than our sparse Gaussian model at 12.8k queries (124 ms, 3.3 GB), while achieving lower accuracy (40.5/30.3 vs. 42.4/31.5).
These results highlight that representing the scene with a compact set of semantic Gaussians, combined with our highly compressed attention module, enables an architecture that is both simple and scalable: it closes the  gap to strong camera-only baselines in terms of efficiency and at the same time clearly outperforms prior multi-modal occupancy methods in the accuracy-efficiency trade-off.

\begin{figure*}[t]
\centering
\includegraphics[width=0.96\linewidth]{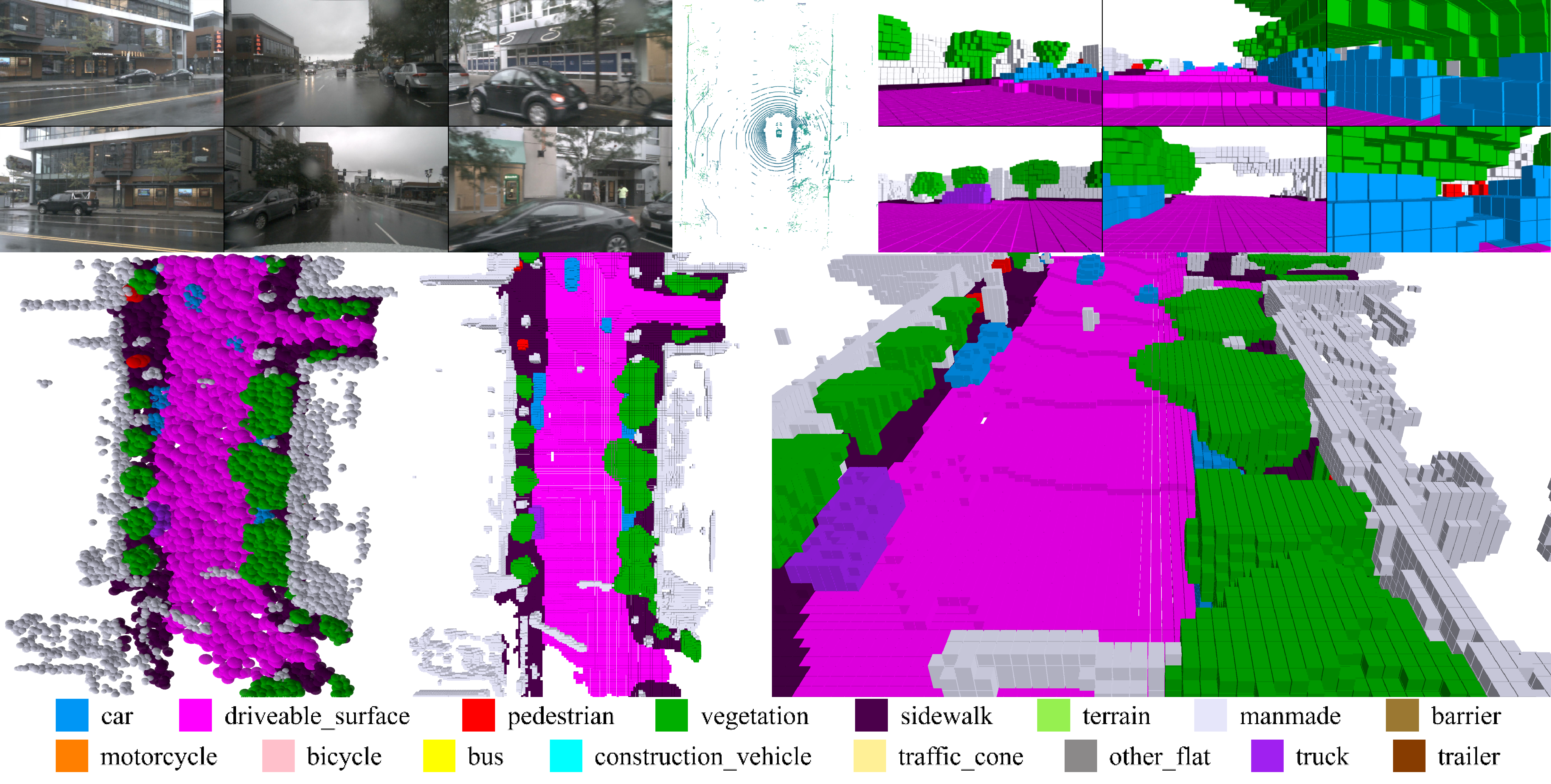} 
\vspace{-0.1cm}
\caption{Additional qualitative results on the SurroundOcc-nuScenes validation set under \textbf{adverse weather} scenarios.\textbf{Top} shows multi-view images (left), raw LiDAR input (center), and predicted image-view occupancy (right); \textbf{Bottom} presents predicted 3D Gaussians, BEV occupancy, and front-view occupancy.}
\label{a1}
\vspace{-0.3cm}
\end{figure*}

\begin{figure*}[t]
\centering
\includegraphics[width=0.96\linewidth]{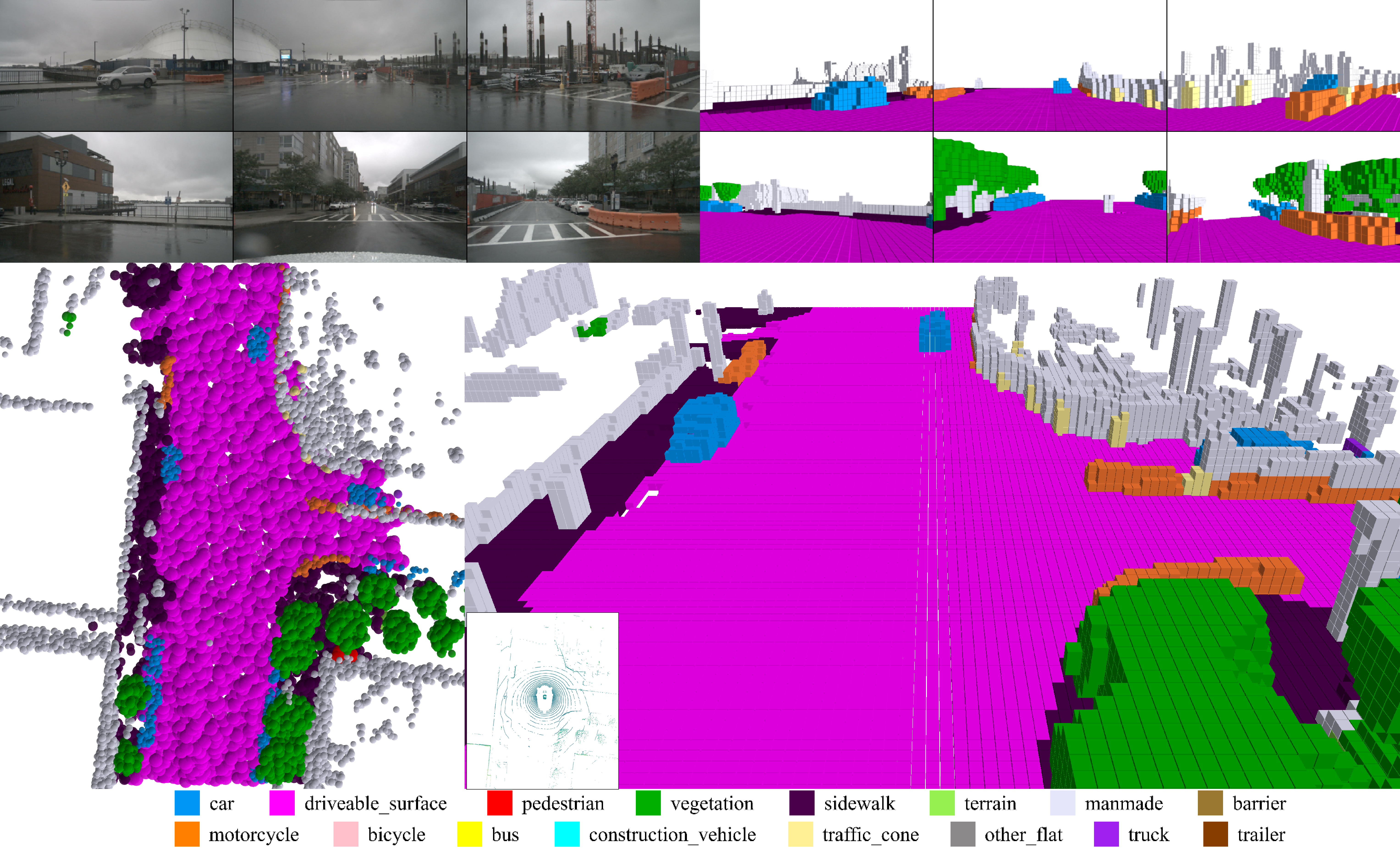} 
\vspace{-0.1cm}
\caption{Additional qualitative results on the Occ3D-nuScenes validation set under \textbf{adverse weather} scenarios.\textbf{Top-left}: multi-view images; \textbf{top-right}: predicted image-view occupancy; \textbf{bottom-left}: predicted 3D Gaussians; \textbf{bottom-right}: front-view occupancy; inset: LiDAR input.}
\label{b1}
\vspace{-0.3cm}
\end{figure*}

\begin{figure*}[t]
\centering
\includegraphics[width=0.96\linewidth]{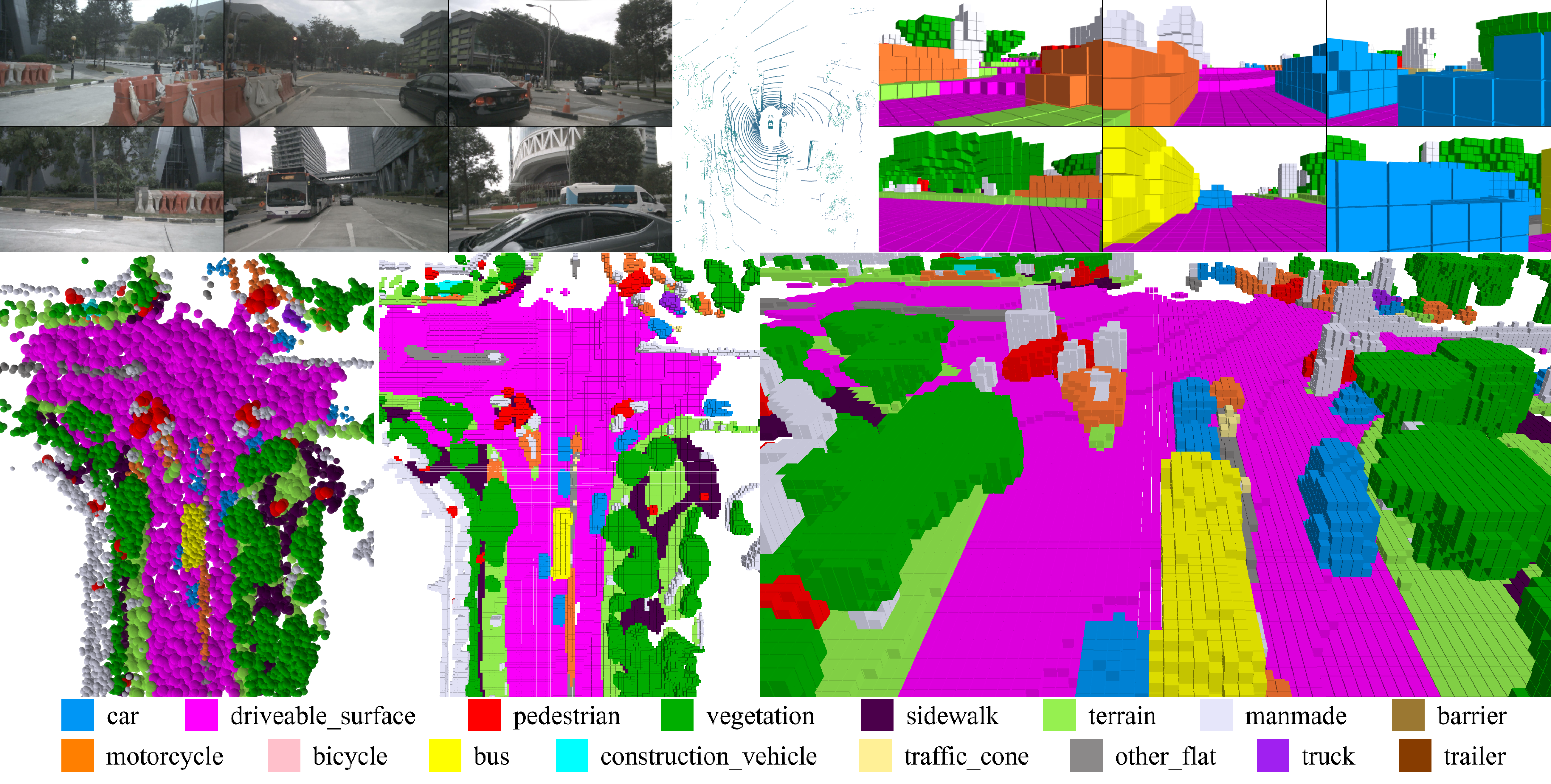} 
\vspace{-0.1cm}
\caption{Additional qualitative results on the SurroundOcc-nuScenes validation set under \textbf{dense traffic} scenarios.}
\label{a2}
\vspace{-0.3cm}
\end{figure*}

\begin{figure*}[t]
\centering
\includegraphics[width=0.96\linewidth]{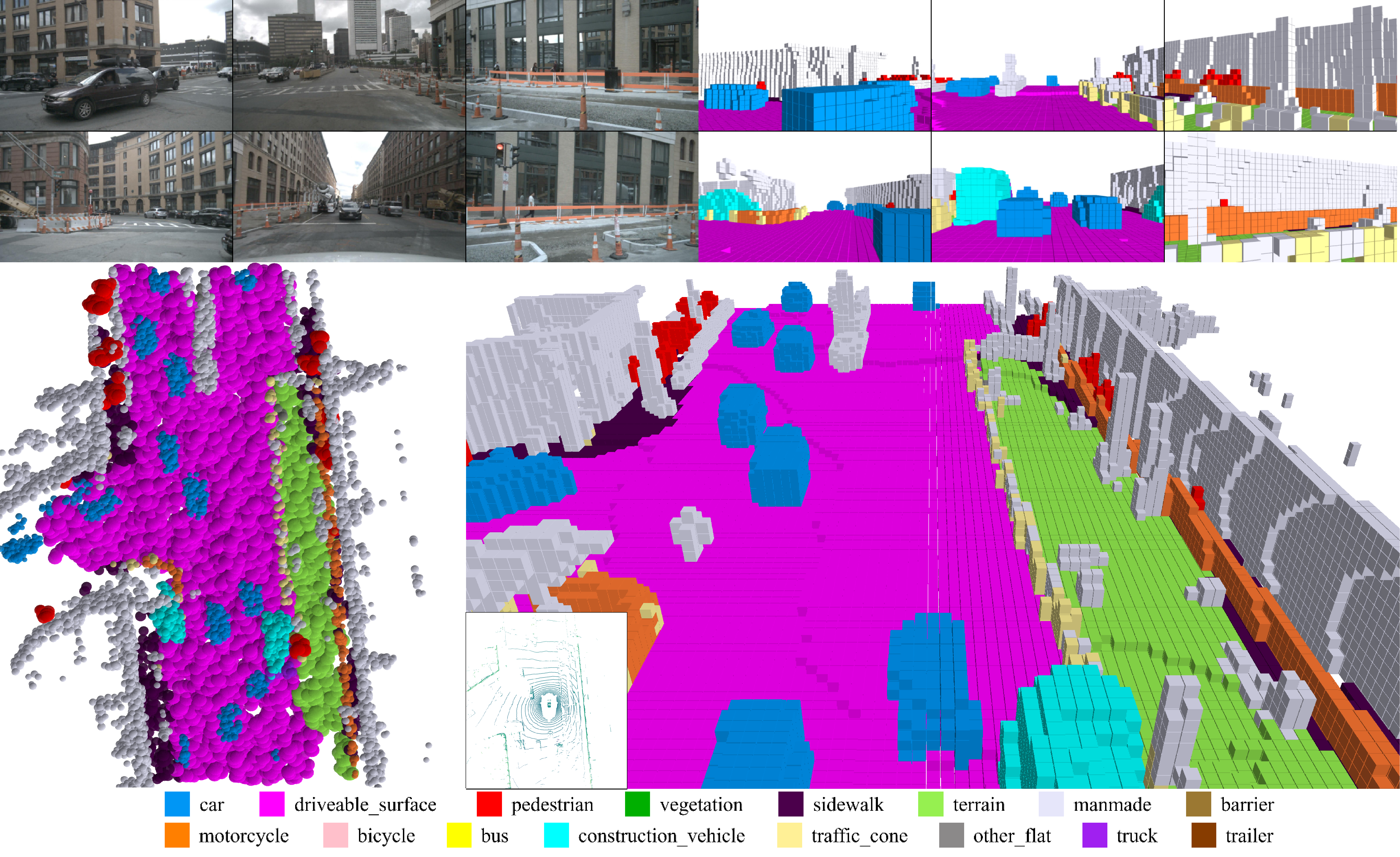} 
\vspace{-0.1cm}
\caption{Additional qualitative results on the Occ3D-nuScenes validation set under \textbf{dense traffic} scenarios.}
\label{b2}
\vspace{-0.3cm}
\end{figure*}

\section{KITTI-360 Result}
In the KITTI-360 benchmark, we provide a comprehensive comparison between \textbf{Gau-Occ} and both LiDAR-only and image-only baselines in Tab.~\ref{tab:comparison_kitti}. Multi-modal methods are scarce on this dataset, so L2COcc~\cite{wang2025l2cocc} serves as the primary strong LiDAR-only reference. As reported in Tab.~\ref{tab:comparison_kitti} (this supplementary material), the proposed Gau-Occ surpasses L2COcc by \textbf{+1.3 IoU} and \textbf{+0.6 mIoU}, while remaining much superior to camera-only methods. Under this challenging single-camera configuration of KITTI-360, our model yields clear gains on moving vehicles (e.g., \textit{car}, \textit{truck}) and large-scale structural classes (e.g., \textit{road}, \textit{building}), indicating improved capability for reliable scene reconstruction from limited visual coverage.

Qualitative results in Fig.~\ref{vis_kitti360} further corroborate these findings: even with a single camera and sparse LiDAR, Gau-Occ reconstructs both global scene layouts and small instances accurately, illustrating robustness to sparse viewpoints and effective exploitation of LiDAR geometry.

\section{Additional Visualizations}

We present additional 3D semantic occupancy prediction results on the Suroundocc-nuScenes and Occ3D-nuScenes validation set. Gau-Occ achieves accurate and complete predictions across diverse challenging scenarios, including \textbf{adverse weather} as shown in Fig.~\ref{a1} and Fig.~\ref{b1} and \textbf{dense traffic} as shown in Fig.~\ref{a2} and Fig.~\ref{b2}.
These results further demonstrate Gau-Occ’s strong generalization and robustness in handling sparse/noisy inputs and reasoning over complex or low-frequency scenes via geometry-aware, multi-modal Gaussian fusion.

\noindent\textbf{Supplementary videos provide dynamic visualizations} of our comparisons with strong baselines, DAOcc and GaussianFormer-2. Our method achieves noticeably higher occupancy accuracy in long-range and heavily occluded regions, and more reliably distinguishes visually similar on-road categories (e.g., \textit{truck} vs.\ \textit{car}) without introducing semantic ambiguity, owing to its more effective use of geometric priors.

\end{document}